\documentclass[times, 5p]{elsarticle}

\usepackage[utf8]{inputenc}
\usepackage{amsmath,amssymb,amstext, bm, mathtools} %
\graphicspath{{figs/}}

\usepackage{tabularray}
\SetTblrInner[tblr]{rowsep=0pt} %
\newcommand{\TCategoryAboveSpace}{\rule{0pt}{15pt}}
\UseTblrLibrary{booktabs}

\usepackage{caption}
\usepackage{subcaption}
\setcitestyle{square}

\DeclareMathOperator*{\argmin}{argmin}

\newcommand\thefont{\expandafter\string\the\font}

\usepackage{xcolor}

\usepackage[hidelinks]{hyperref}
\hypersetup{pdfauthor=author}
\usepackage{cleveref}

\newcommand{\doi}[1]{doi: \url{#1}}

\begin{document}

\begin{frontmatter}

\title{\bf Recovery Policies for Safe Exploration of Lunar\\ Permanently Shadowed Regions by a Solar-Powered Rover}

\author[1]{Olivier Lamarre\corref{cor1}\fnref{fn1}}
\author[2]{Shantanu Malhotra\fnref{fn2}}
\author[1]{Jonathan Kelly\fnref{fn3}}

\cortext[cor1]{Corresponding author}
\fntext[fn1]{Email: {\tt\footnotesize olivier.lamarre@robotics.utias.utoronto.ca}}
\fntext[fn2]{Email: {\tt\footnotesize shantanu.malhotra@jpl.nasa.gov}}
\fntext[fn3]{Email: {\tt\footnotesize jonathan.kelly@robotics.utias.utoronto.ca}}

\affiliation[1]{
    organization={Space \& Terrestrial Autonomous Robotic Systems (STARS) Laboratory\\ University of Toronto Institute for Aerospace Studies},
    addressline={4925 Dufferin Street},
    city={Toronto, Ontario},
    postcode={M3H 5T6},
    country={Canada}
}
\affiliation[2]{
    organization={Jet Propulsion Laboratory\\ California Institute of Technology},
    addressline={4800 Oak Grove Drive},
    city={Pasadena, California},
    postcode={91109},
    country={United States}
}

\begin{abstract}
The success of a multi-kilometre drive by a solar-powered rover at the lunar south pole depends upon careful planning in space and time due to highly dynamic solar illumination conditions.
An additional challenge is that the rover may be subject to random faults that can temporarily delay long-range traverses.
The majority of existing global spatiotemporal planners assume a deterministic rover-environment model and do not account for random faults.
In this paper, we consider a random fault profile with a known, average spatial fault rate. 
We introduce a methodology to compute recovery policies that maximize the probability of survival of a solar-powered rover from different start states.
A recovery policy defines a set of recourse actions to reach a safe location with sufficient battery energy remaining, given the local solar illumination conditions.
We solve a stochastic reach-avoid problem using dynamic programming to find an optimal recovery policy.
Our focus, in part, is on the implications of state space discretization, which is required in practical implementations.
We propose a modified dynamic programming algorithm that conservatively accounts for approximation errors.
To demonstrate the benefits of our approach, we compare against existing methods in scenarios where a solar-powered rover seeks to safely exit from permanently shadowed regions in the Cabeus area at the lunar south pole.
We also highlight the relevance of our methodology for mission formulation and trade safety analysis by comparing different rover mobility models in simulated recovery drives from the LCROSS impact region.
\end{abstract}


\end{frontmatter}

\section{Introduction}

The accumulation of water and other volatiles in permanently shadowed regions (PSRs) at the lunar poles has been hypothesized for decades \cite{watson_behavior_1961}.
In 2018, a new analysis of the Moon Mineralogy Mapper instrument data confirmed the presence of surface water ice in PSRs, the majority of which is located at the lunar south pole \cite{li_direct_2018}.
This discovery has motivated a rise in long-range surface robotic mission case studies investigating key prospecting areas \cite{flahaut_regions_2020, lemelin_framework_2021, losekamm_assessing_2022}.
Such missions would reveal new information about the formation of our solar system and inform the location of resources to support a sustained human presence on the lunar surface.
Scheduled to launch in 2024, NASA's solar-powered Volatiles Investigating Polar Exploration Rover (VIPER) will be the first surface mission to focus on these objectives \cite{nasa_viper_2020}.

A central challenge for solar-powered exploration of the lunar south pole is the terrain topography: large and fast-moving shadows are cast due to the low sun elevation above the horizon.
While insolation conditions can vary dramatically from one region to the next during the lunar synodic day (approximately 29.5 Earth days \cite{heiken_lunar_1991}), a solar-powered rover requires, at the very least, intermittent sun exposure to maintain sufficient energy levels throughout its mission.
As such, planning in space and time (i.e., spatiotemporally) is crucial to ensure rover safety \cite{teti_sun-synchronous_2005}.

\begin{figure*}[t]
    \centering
    \includegraphics[width=0.96\textwidth]{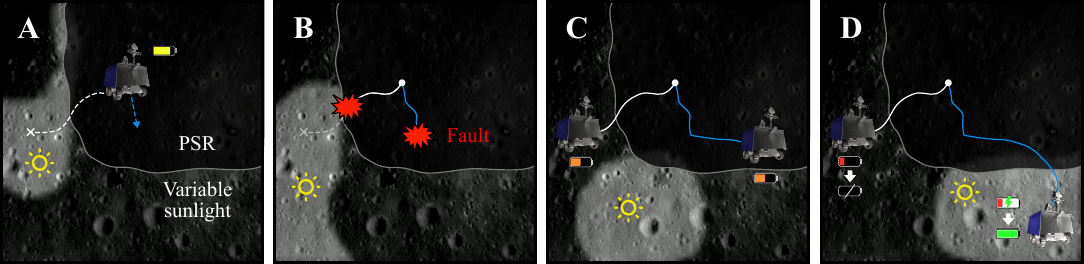}
    \caption{Impact of faults on the safety of a solar-powered rover exiting a PSR. In subfigure A, the dashed line indicates the path generated by a hypothetical risk-agnostic offline spatiotemporal planner. The blue arrow shows an action returned by a hypothetical online, risk-aware planner.
    A fault occurring early in the traverse (subfigure B) not only invalidates the offline plan, but in this case also prevents the rover from being exposed to sunlight, no matter where it exits the PSR (subfigure C).
    On the other hand, a risk-aware online planner can, by design, proactively account for stochastic faults (see blue path).
    In this conceptual example, the online planning methodology leads the rover to sunlight (subfigure D).
    Background image courtesy of NASA and Arizona State University.
    VIPER render courtesy of NASA.}
    \label{fig:overview}
    \vspace{-1.5mm}
\end{figure*}

A common assumption in work on spatiotemporal planning for planetary environments is that the rover-environment interaction model is fully known.
In reality, as demonstrated over the past several decades by the robotic Martian missions, faults can happen unexpectedly and delay traverse progress~\cite{rankin_mars_2021}.
Furthermore, with the extreme insolation conditions at the lunar south pole, some delays may have a greater impact on a solar-powered rover than others, depending on \textit{when} and \textit{where} they occur.
For instance, an unforeseen delay of a few hours inside of a PSR might, at best, force the rover to exit the PSR early or at a different location than initially intended.
At worst, such a delay may cause the rover to miss future periods of sun exposure that are crucial to its survival.
As such, understanding the risks associated with any spatiotemporal global rover state is necessary to ensure safe long-term mobility.
We illustrate the situation graphically in~\Cref{fig:overview}.

In this paper, we build upon existing work in the area of reachability analysis, that is, determining whether a controlled system is capable of reaching a certain region of the state space.
We quantify the risk associated with the exploration of PSRs at the lunar south pole by a solar-powered rover affected by random faults that cause navigation delays.
By `risk', we mean the probability that the rover fails to reach the designated, safe region of the state space.
In practice, a navigation delay could represent the time necessary to resolve an issue on board the rover, or, with support from operations teams on Earth, before resuming navigation.
Our work is therefore relevant to mission planning and formulation, long-term spacecraft autonomy, and ground-in-the-loop (GIL) mission assistance.

We investigate the problem of finding a recovery policy that a solar-powered rover can follow to maximize the probability of reaching a safe region from any feasible initial state.
Informed by prior, empirical studies indicating that faults occur at a known average rate\footnote{In~\cite{robinson_intrepid_2020,keane_endurance_2022}, the average distance between faults is derived from previous Mars mission data.}~\cite{robinson_intrepid_2020,keane_endurance_2022,thangeda_adaptive_2022},
we assume that the number of faults over a given drive distance follows a Poisson distribution, which has unbounded support.
As such, there is always a non-zero, albeit small, probability of experiencing a series of faults that violate any deterministic safety threshold.

Our global mobility model is formalized as a controlled discrete time stochastic hybrid system (DTSHS).
Such systems have dynamics defined across a state space that contains both discrete and continuous components. %
In the context of planetary mobility, continuous state variables may include the time of day, subsystem temperatures, or battery energy level, while discrete components might represent operational modes, rover health, or any other variable that has a countably finite set of possible values.
In practice, most numerical solutions to DTSHS reachability problems employ a discretization of the continuous components of the state space without considering the impact that such an approximation can have on the safety of the underlying (real-world) system.
In this work, we employ a min-max dynamic programming paradigm to conservatively account for undesirable discretization effects on risk predictions.
The contributions of our work are fourfold:
\begin{enumerate}
    \item We formulate a DTSHS reach-avoid problem to determine maximally safe policies tailored for solar-powered exploration of the lunar south pole.
    \item We implement a min-max dynamic programming scheme to mitigate undesirable, safety-compromising effects associated with standard discretization methods.
    \item We empirically validate our approach on real orbital terrain and illumination maps of the lunar south pole, simulating safe exits from PSRs in the Cabeus region.
    \item We release \texttt{gplanetary\_nav}, an open-source Python library to preprocess orbital data in support of kilometre-scale traversability planning in planetary environments.
\end{enumerate}
The remainder of the paper is structured as follows.
\Cref{sec:relatedwork} reviews prior work on sun-synchronous navigation for solar-powered field robots. %
We also provide an overview of relevant stochastic reachability results from the field of safe optimal control.
In \Cref{sec:problemstatement}, we instantiate our global solar-powered rover mobility model as a DTSHS and formalize the stochastic reach-avoid problem that is central to the paper.
\Cref{sec:planning_approach} describes our dynamic programming approach to mitigate risk mischaracterizations caused by discretization.
We empirically compare our approach against two existing approximation methods in \Cref{sec:experiments}, by simulating safe navigation recovery drives from large PSRs at the lunar south pole.
Lastly, in \Cref{sec:exp3}, we present a practical example involving simulated drives from the LCROSS impact area.\footnote{Interested readers can find additional results at \url{https://papers.starslab.ca/recovery-policies-psr-exploration}.}

\section{Related Work}
\label{sec:relatedwork}

We begin by briefly reviewing various efforts in the area of spatiotemporal planning for long-range solar-powered surface mobility.
These approaches mostly assume a deterministic rover-environment model.
We then discuss results in stochastic reachability analysis that are important to the approach presented in this paper.

\subsection{Spatiotemporal Global Navigation Planning}

Spatial and temporal planning are essential for successful sun-synchronous planetary exploration, where the goal is to maximize exposure to sunlight \cite{whittaker_sun-synchronous_2000}.
Sun-synchronous mobility was initially field-demonstrated in the Canadian Arctic by the Temporal Mission Planner for the Exploration of Shadowed Terrain (TEMPEST). TEMPEST is a global planner that tightly couples path selection and resource management (in this case, energy)~\cite{wettergreen_sun-synchronous_2005}.
The work on TEMPEST eventually inspired the development of planning techniques for the solar-powered exploration of the lunar south pole.
The authors of~\cite{otten_planning_2015} use connected component analysis in the spatiotemporal domain to generate routes that are constantly illuminated.
This approach is extended to account for Earth communication blackouts in~\cite{otten_strategic_2018}.
The method in~\cite{cunningham_accelerating_2017} relies on heuristics to accelerate the performance of an energy-aware A* planner. 

Resource-constrained navigation planning under random disturbance is studied and formalized as a consumption Markov decision process (CMDP) in~\cite{blahoudek_qualitative_2020,blahoudek_efficient_2023}.
The CMDP framework defines dynamics that allow an agent to recharge a resource by taking a single action in special ``reload'' states.
However, it is unclear how this framework could be leveraged to address solar-powered mobility problems; the ability of a rover to recharge its batteries depends on where it is, at what time, and under what illumination conditions.
In the context of global planning for the lunar surface, the body of work accounting for stochastic rover-environment interactions is still very small.
For instance, the method in~\cite{hu_planning_2022} generates long-range plans using custom, heuristic safety functions.
In \cite{inoue_spatio-temporal_2021}, the cost function of a spatiotemporal planner incorporates an error term dependent on whether the state is in a (time-varying) shadow area and/or in a communication blackout area.
The components of the error are weighted by the probabilities of experiencing a number of delays during a transition. %
A limitation of this work is the assumption that the rover must remain in sunlight at all times---in contrast, we are interested in the exploration of permanently shadowed regions.
Risk- and resource-aware long-term spatiotemporal planning for the solar-powered exploration of dynamically-lit environments (like the lunar south pole) remains an open challenge.

\subsection{Stochastic Reachability Analysis for DTSHS}

Reachability analysis for stochastic hybrid systems primarily relies on dynamic programming (DP) to determine the risk associated with different policies and initial start states.
Abate et al.~\cite{abate_probabilistic_2008} investigate reachability in the discrete time case and define safety as the ability of a (controlled) system to remain inside a safe/target region of the state space over a finite time horizon.
The authors formulate reachability as a stochastic optimal control problem, where the solution is a Markov policy that maximizes a multiplicative value function.
This methodology is extended to reach-avoid problems in \cite{summers_verification_2010}.
A reach-avoid problem involves controlling a DTSHS to maximize the probability of hitting a target region of the state space while avoiding an undesirable/dangerous set of states.
One interpretation of this problem, the ``first hitting time'' case, requires finding a Markov policy that maximizes a sum-multiplicative value function.
As explained below, this framework influences our approach, since it is crucial for a solar-powered planetary rover to stay within reach of sunlight and avoid operational failure states (such as running out of energy or falling behind schedule).

Reachability analysis of a DTSHS has also been studied from a dynamic game-theoretic perspective.
The work in~\cite{ding_stochastic_2013} focuses on max-min stochastic reachability in the context of a zero-sum game where an additional (adversarial) agent acts in a way to minimize the probability of safety.
In~\cite{yang_dynamic_2018}, a robust approach is presented which maximizes the probability that a stochastic system remains in a safe region over a finite time horizon when the disturbance distribution is unknown.

In the optimal control literature, the above approaches are often considered ``risk-neutral'' because they only capture the probability of undesirable outcomes, as opposed to their severity.
For instance,~\cite{chapman_risk-sensitive_2019} proposes a risk-sensitive reachability method that incorporates a risk measure to control the frequency of constraint violation in addition to mitigating the severity of violations.
Risk-sensitive reachability falls outside the scope of this paper; we assume that entering a dangerous region of the state space will result in a mission-ending scenario.
In turn, constraint violation severity is irrelevant because there are no recourse actions available once in a failure state.
We refer the reader to the survey in~\cite{wang_risk-averse_2022} of related risk-neutral and risk-averse optimal control methods for more details.

One focus of our investigation is the way in which the reachability analyses reviewed so far are applied to the real world.
These analyses use numerical approximations to remain computationally tractable.  We pay attention to the effects that approximations have on the reliability of risk predictions.
All the methods described to this point discretize the continuous dimensions of the state space.
For instance, the work in \cite{abate_probabilistic_2008} turns the continuous components of the state space into a grid and maps all hybrid states within the same grid partition to a single discrete reference state, akin to a nearest neighbour paradigm \cite{abate_computational_2007}.
The approaches in \cite{summers_verification_2010} and \cite{ding_stochastic_2013} follow a similar approximation methodology.
Unfortunately, neither \cite{summers_verification_2010} nor \cite{ding_stochastic_2013} validate their numerical approximations through simulations of the underlying hybrid system.
In~\cite{chapman_risk-sensitive_2019}, on the other hand, multi-linear interpolation between neighbouring grid points is employed and Monte Carlo experiments are carried out to validate the procedure.
Herein, we demonstrate how a nearest neighbour or multi-linear interpolation paradigm can lead to optimistic risk predictions in the context of planetary rover navigation when the robot is affected by random delays.
We instead rely on a min-max formulation to conservatively account for errors caused by the discretization of the state space.

\section{Problem Statement}
\label{sec:problemstatement}

We begin by describing our rover mobility model as a specific DTSHS instance.
This instance is a simplified version of the one introduced in~\cite{abate_probabilistic_2008}, since we only define the hybrid state space, action space, and a single stochastic transition function acting over the entire state space (as opposed to distinct stochastic kernels for different state transitions).
Our notation roughly follows that in ~\cite{abate_probabilistic_2008}.
We employ state and action spaces inspired by previous spatiotemporal planners (such as~\cite{wettergreen_sun-synchronous_2005,cunningham_accelerating_2017}) but with nondeterministic dynamics.
Then, we introduce the notion of safe states for solar-powered mobility, which act as members of the target set for our particular stochastic reach-avoid problem.

\subsection{Rover Mobility Model as a DTSHS}
Consider a hybrid state space whose elements are described by the 3-tuple $\left(c,\, t,\, b\right)$, where $c \in \mathcal{C} \subset \mathbb{N} \times \mathbb{N}$ is a cell on a discrete grid representing possible physical rover locations on the lunar surface.
In this work, grid cells in $\mathcal{C}$ are mapped to pixels on an orbital map.
Variable $t \in \mathbb{R}_{\geq\,0}$ is the time and $b \in \mathbb{R}_{\geq\,0}$ is the rover's battery energy level, also referred to as the state of charge (SOC).
Here, the (continuous) time component of our hybrid state space is not related to the  term ``discrete time'' (DT) in DTSHS.
The latter simply refers to the discrete steps taken in succession to move through the state space.
We assume that $\mathcal{C}$ only contains locations where the terrain type and geometry allow safe mobility.

The discrete state subspace is the set of locations $\mathcal{C}$ and the continuous state subspace associated with each location is $\mathbb{R}_{\geq\,0} \times \mathbb{R}_{\geq\,0}$.
Our hybrid state space is then ${\mathcal{X} \coloneqq \mathcal{C} \times \mathbb{R}_{\geq\,0} \times \mathbb{R}_{\geq\,0}}$.
With a slight abuse of notation, let the choice functions $c(\boldsymbol{x})$, $t(\boldsymbol{x})$, and $b(\boldsymbol{x})$ define, respectively, the grid cell, time, and battery energy associated with state ${\boldsymbol{x} \in \mathcal{X}}$.

The action space $\mathcal{A}$ defines a finite set of mobility actions.
From any state, nine mobility actions are possible, including driving to one of the neighbouring grid cells in $\mathcal{C}$ (eight possible actions) or waiting in place for the fixed duration $\delta t_{\mathrm{wait}}$.
When the rover waits in place, only the change in SOC depends on the local environmental (insolation) conditions.
If the rover instead drives to one of the eight neighbouring grid cells, both the action duration and the change in SOC are dependent on the local conditions (terrain type, geometry, and insolation).
We do not track the rover's azimuth and instead assume that the power generated is approximately the same for all headings.
This is a reasonable assumption if the rover is equipped with zenith-facing solar panels or, more appropriately for polar mobility, tilted panels that are evenly distributed on all its sides or an actuated array platform.

Our system dynamics are affected by random faults that may halt the rover during the drive to a neighbouring grid cell.
Drawing inspiration from \cite{robinson_intrepid_2020, keane_endurance_2022, thangeda_adaptive_2022}, we assume a known average spatial fault rate $\alpha$ and a fixed fault recovery period $\delta t_{\mathrm{fault}}$ during which the rover remains in place until the fault is resolved.
We also assume that the fault probabilities over disjoint drive intervals are independent. 
In turn, the fault profile is a Poisson process~\cite{ross_first_2019}, which implies that the (spatial) distance between consecutive faults follows an exponential distribution.
Let the Poisson random variable $F$ define the number of faults that the rover experiences over a given driving distance.
If $\rho(c,a)$ represents the driving distance when taking mobility action $a$ from grid cell $c$, the probabilities of zero or one or more faults occurring are, respectively,
\begin{equation}
\begin{aligned}
    \mathrm{Pr}(F = 0; c,a)    & =     \exp\big(\!\left.-\alpha\rho(c,a)\right.\!\big),\\
    \mathrm{Pr}(F \geq 1; c,a) & =  1 -\exp\big(\!\left.-\alpha\rho(c,a)\right.\!\big).    
\end{aligned}
\label{eq:fault_probs}
\end{equation}
Let the function $\rho: \mathcal{C} \times \mathcal{A} \rightarrow \mathbb{R}_{\geq\,0}$ map to the Euclidean driving distance between grid cell centres in three dimensions, based on orbital elevation data, or to zero (if the `wait in place' action is taken).
Only driving actions (which are associated with positive driving distances) are subject to faults; the action of waiting in place always terminates nominally.
We assume that the state is fully observable at all times.

\begin{figure}[t]
    \centering
    \includegraphics[width=\columnwidth]{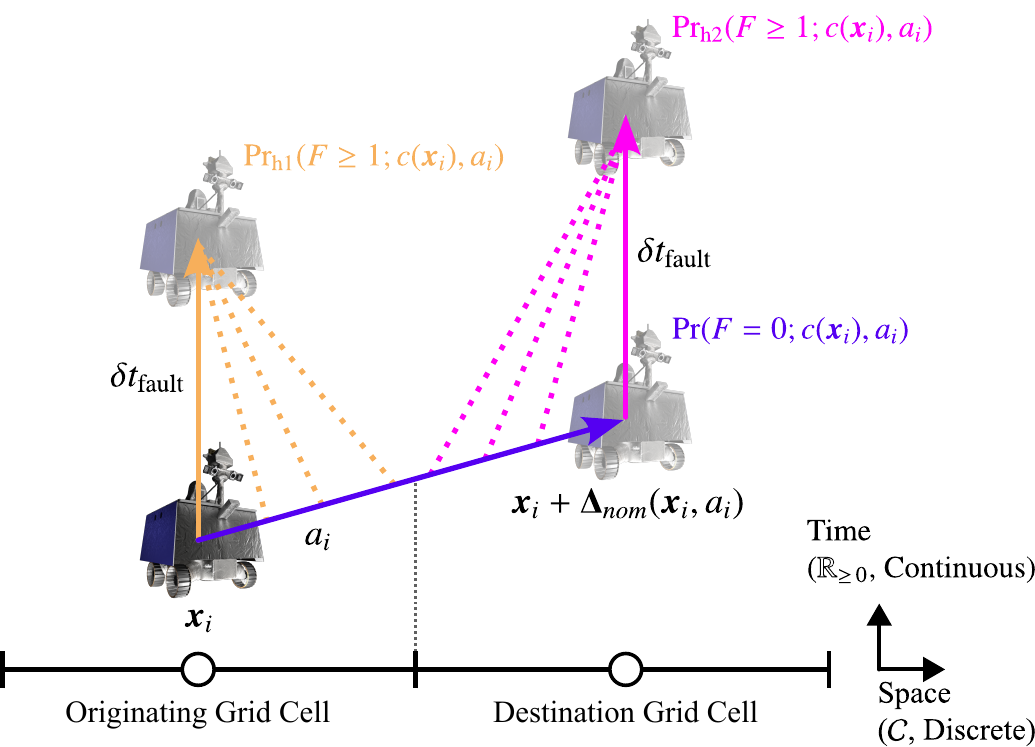}
    \caption{Spatiotemporal view of possible state transition outcomes when driving to a neighbouring grid cell (energy dimension not shown): a nominal transition (no faults), a fault occurring during the first half of the drive, and a fault only occurring during the second half.}
    \label{fig:transitions}
\end{figure}

\Cref{fig:transitions} illustrates the different ways that faults are accounted for in the hybrid state space.
For a single action (a single step in the state space), we define the corresponding state transition by accounting for when the \textit{next} fault might occur only.\footnote{This is equivalent to the probability of at least one fault occurring over a given drive interval, as illustrated in~\Cref{eq:fault_probs}. We maintain this notation for mathematical correctness and consistency.}
Upon taking drive action $a$ from grid cell $c$ to a neighbouring location, a nominal transition occurs with probability $\mathrm{Pr}(F = 0; c,a)$ as defined by~\Cref{eq:fault_probs}.
When considering transitions affected by a fault, we divide the drive into two halves to (approximately) model the transition from the originating grid cell to the destination grid cell.
The probability that the next fault occurs anywhere in the first half of the traverse is
\begin{equation}
    \text{Pr}_{\text{h1}}(F \geq 1; c,a) = 1 - \exp\big(-\alpha\rho(c,a)/2\big).
    \label{eq:half1_prob}
\end{equation}
The corresponding transition is approximated by a forced delay of duration $\delta t_{\mathrm{fault}}$ from the start state.
The probability that the next fault occurs in the second half of the traverse (rather than the first) is
\begin{equation}
    \text{Pr}_{\text{h2}}(F \geq 1; c,a) = \exp\big(-\alpha\rho(c,a)/2\big)-\exp\big(-\alpha\rho(c,a)\big),
    \label{eq:half2_prob}
\end{equation}
which is approximated by a forced delay of duration $\delta t_{\mathrm{fault}}$ from when the nominal transition would have ended.

The stochastic state transition dynamics are defined by
\begin{equation}
\begin{aligned}
    \boldsymbol{x}_{i+1} = & \quad \boldsymbol{f}(\boldsymbol{x}_i, a_i) \\
    = & \quad \boldsymbol{x}_i + \boldsymbol{\Delta}_{\text{sto}}(\boldsymbol{x}_i, a_i),
\end{aligned}
\end{equation}
where subscript $i$ specifies the current planning step index and $\boldsymbol{\Delta}_{\text{sto}}(\boldsymbol{x}_i, a_i) \sim \mathcal{D}_{\text{sto}}(\boldsymbol{x}_i, a_i)$ is a random variable representing the change in state parameters when taking action $a_i$ from state $\boldsymbol{x}_i$.
The set $\mathcal{D}_{\text{sto}}(\boldsymbol{x}_i, a_i)$ includes the three possible realizations for $\boldsymbol{\Delta}_{\text{sto}}(\boldsymbol{x}_i, a_i)$.
These realizations correspond to the three state transitions shown in \Cref{fig:transitions} and the respective probabilities defined by \Cref{eq:fault_probs,eq:half1_prob,eq:half2_prob}:
\begin{multline}
    \mathcal{D}_{\text{sto}}(\boldsymbol{x}_i, a_i) = \\
    \begin{cases}
        \boldsymbol{\Delta}_{\text{nom}}(\boldsymbol{x}_i, a_i) & \text{prob. }\text{Pr}(F = 0; c(\boldsymbol{x}_i),a_i), \\[5pt]
        \boldsymbol{\Delta}_{\text{fault}}(\boldsymbol{x}_i) & \text{prob. }\text{Pr}_{\text{h1}}(F \geq 1; c(\boldsymbol{x}_i),a_i), \\[5pt]
    \!\begin{aligned}
        & \boldsymbol{\Delta}_{\text{nom}}(\boldsymbol{x}_i, a_i) + \\
        & \boldsymbol{\Delta}_{\text{fault}}(\boldsymbol{x}_i+\boldsymbol{\Delta}_{\text{nom}}(\boldsymbol{x}_i, a_i)) 
    \end{aligned} & \text{prob. }\text{Pr}_{\text{h2}}(F \geq 1; c(\boldsymbol{x}_i),a_i), \\
    \end{cases}
\end{multline}
where $\boldsymbol{\Delta}_{\text{nom}}(\boldsymbol{x}_i, a_i)$ denotes the change in state parameters when a nominal (fault-free) transition occurs.
The function $\boldsymbol{\Delta}_{\text{fault}}(\boldsymbol{x}_i)$ returns the change in state parameters when a fault occurs and the rover stays in place for a duration of $\delta t_{\text{fault}}$ starting from state $\boldsymbol{x}_i$.
The independence of fault probabilities between successive mobility steps (which take place over disjoint drive intervals) is determined by the modelling of fault profiles as Poisson processes.
In this work, we focus on a constant fault recovery time for simplicity.
However, our approach can accommodate discretely-distributed random recovery times.
This would multiply the number of possible outcomes when taking an action from a specific state.

\subsection{Safe Havens and Safe States for Solar-Powered Mobility}
\label{sec:safehaven}

Generally speaking, safe havens are locations that receive sufficient sunlight to keep a solar-powered rover `alive' over a certain period of time (these ares are informally called `lily pads' in the context of Martian exploration \cite{webster_nasa_2017}).
Safe havens are useful from a safety standpoint because they are areas where a solar-powered rover can stay, for some period of time, with limited to no human operator oversight.
Likewise, safe havens also help with the creation of long-term traverse plans across regions with challenging insolation conditions.

For site selection and preliminary traverse planning at the lunar south pole, the VIPER mission defines a safe haven as any location receiving sunlight for a minimum duration while the Earth is below the horizon \cite{shirley_overview_2022}.
However, characterization based on a minimum insolation duration is ineffective for more detailed traverse planning in two ways:
\begin{enumerate}
    \item Alone, reaching a safe haven is insufficient to guarantee the survival of a (solar-powered) rover. For instance, if the rover happens to arrive with a low SOC at the beginning of a shadow period, full battery depletion might be unavoidable. A safety criterion dependent on the rover's arrival time and SOC at different safe havens is necessary.
    \item Relying on insolation duration at safe havens can be very constraining. For instance, in~\cite{shirley_overview_2022} numerous regions of interest at the lunar south pole are not even considered.\footnote{The VIPER landing site selection also depends on other factors like direct-to-Earth communications, but insolation time while the Earth is below the horizon remains an important contributing element.}
    Similar accessibility criteria are used in support of preliminary landing site selection analyses for the Artemis program in~\cite{brown_resource_2022}.
\end{enumerate}

We overcome these challenges with a more complete and permissive definition of a safe haven.
Fundamentally, a safe haven can be any location on the lunar surface outside of permanently shadowed regions.
This set of locations can be further constrained based on mission-specific requirements.
An acceptable traverse profile should end up in some safe haven ${h \in \mathcal{C}}$ where a minimum energy threshold $\bar{b}_h$ must be satisfied within a (user-defined) time limit $\bar{t}_h$.
From an operational perspective, imposing such boundary conditions could be useful to ensure the rover meets some mission requirement.
Examples include participating in an orbital communication pass, rendezvousing with a lander that must respect a tight launch schedule, or simply being ready for the next leg of a long traverse.
From a safe optimal control perspective, it allows the definition of a bounded set of safe states ${\mathcal{S} \subset \mathcal{X}}$ that form the target set.

\begin{figure}[t]
    \centering
    \includegraphics[width=\columnwidth]{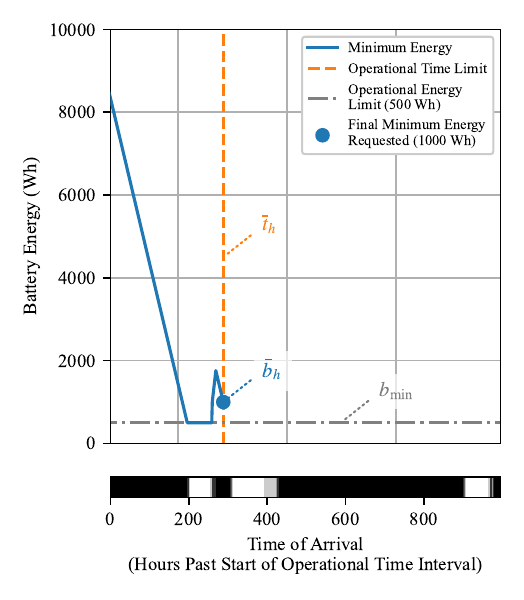}%
    \vspace*{-3mm}
    \caption{Example of the minimum required rover SOC as a function of arrival time at a safe haven. A rover at this location with a SOC above the curve is considered to be `safe.' The horizontal grayscale bar at the bottom qualitatively shows the insolation conditions at the location over time (white represents full insolation, black is full shade).}
    \label{fig:safehaven}
\end{figure}

For every safe haven, we use the boundary conditions defined above and iterate backwards in time to calculate $\zeta_h(t)$, a minimum rover energy time series, using the rover energy model and solar illumination data (see \Cref{fig:safehaven} for an example).
Intuitively, this map defines the minimum SOC that the rover needs in order to hibernate at $h$ upon arriving at some time $t \leq \bar{t}_h$ (and to ensure that the energy boundary condition is satisfied at $\bar{t}_h$).
Thus, the set of safe states is defined as
\begin{equation}
    \mathcal{S} = \{\boldsymbol{x} \in \mathcal{X} \; | \; \zeta_{c(\boldsymbol{x})}(t(\boldsymbol{x})) \leq b(\boldsymbol{x}) \},
\end{equation}
where $c(\boldsymbol{x})$, $t(\boldsymbol{x})$ and $b(\boldsymbol{x})$ are choice functions described previously.
By definition, ${\zeta_{h}(t) = \infty \; \forall t > \bar{t}_h}$.

\subsection{Optimization Formalization}

Let $\mathcal{T}=[t_{\mathrm{min}},t_{\mathrm{max}}]$ be a valid time interval from an operational standpoint (where $t_{\mathrm{min}}$ represents the earliest time during which the rover could be driving and $t_{\mathrm{max}}$ is the latest time limit associated with all safe havens).
Similarly, let $\mathcal{B} = [b_{\mathrm{min}},b_{\mathrm{max}}]$ represent the interval of operational rover energy levels (a logical choice for $b_{\mathrm{max}}$ would be the rover's battery capacity).
The hybrid set $\mathcal{O} \coloneqq \mathcal{C} \times \mathcal{T} \times \mathcal{B}$, of which the safe set $S$ is a subset, represents the operational region of the global state space $\mathcal{X}$.
We assume that stepping outside of $\mathcal{O}$ always results in a mission failure.

\begin{figure}[t]
    \centering
    \includegraphics[width=0.7\columnwidth]{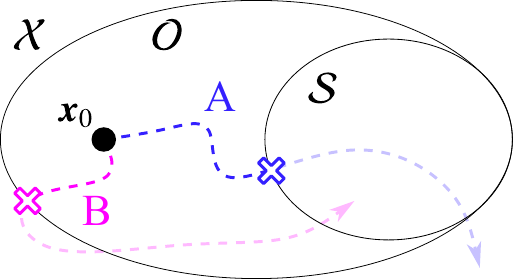}
    \caption{The relation of the operational region $\mathcal{O}$ and the safe set $\mathcal{S}$ within the global hybrid state space $\mathcal{X}$. The cost of trajectories A and B (based on \Cref{eq:cost}) is 0 and 1, respectively. Intuitively, the cost solely depends on whether the trajectory first enters the safe or failure region (these first entry points are indicated by an X). This is the ``first hitting time'' case presented in~\cite{summers_verification_2010}. Elongating trajectories beyond this transition point (illustrated above with semi-transparent segments) does not alter their cost. In practice, valid trajectories would not extend beyond terminal states.}
    \label{fig:cost-example}
\end{figure}

Let $\{\boldsymbol{x}_0, \boldsymbol{x}_1, ..., \boldsymbol{x}_{N-1}\}$ denote a trajectory (a sequence of $N$ consecutive states at different discrete stage indices) sufficiently long such that either the safe or failure set was entered once.
With minor change to sum-multiplicative formula presented in \cite[Section 3.1]{summers_verification_2010}, we also denote the cost associated with a state trajectory using a sum-multiplicative formula:
\begin{multline}
    \sum^{N-1}_{j=0} \left( \prod^{j-1}_{i=0} \boldsymbol{1}_{\mathcal{O}\setminus\mathcal{S}}(\boldsymbol{x}_i) \right) \boldsymbol{1}_{\mathcal{X}\setminus\mathcal{O}}(\boldsymbol{x}_j) \\
    =\begin{cases}
    0 & \text{if the trajectory enters $\mathcal{S}$ before entering $\mathcal{X}\setminus\mathcal{O}$}\\
    1 & \text{otherwise},
    \end{cases}
    \label{eq:cost}
\end{multline}
where $\boldsymbol{1}_{A}(\cdot): \mathcal{X} \rightarrow \{0,1\}$ is the indicator function for a set $A$ and the edge case
$\prod^{-1}_{i=0}(\cdot)=1$ by convention (occuring when $j=0$).
Refer to \Cref{fig:cost-example} for an example.

We seek an online control law (a deterministic Markov policy $\pi: \mathcal{X} \rightarrow \mathcal{A}$) that maximizes the probability of reaching a safe state $\boldsymbol{x} \in \mathcal{S}$ before entering the failure region $\mathcal{X}\setminus\mathcal{O}$.
We formulate finding such a control law as a reach while avoiding problem (specifically, the ``first hitting time'' case) inspired by the work on the topic in~\cite{summers_verification_2010}.
For any initial state $\boldsymbol{x}_0 \in \mathcal{X}$, an optimal policy $\pi^*$ is defined as follows:
\begin{equation}
    \pi^* = \argmin_{\pi \in \Pi} r^\pi_{\boldsymbol{x}_0}(\mathcal{S},\mathcal{O}),
\end{equation}
where $\Pi$ is the set of Markov policies and $r^\pi_{\boldsymbol{x}_0}(\mathcal{S},\mathcal{O})$ is the \textit{risk} (probability of \textit{not} reaching the safe set) associated with policy $\pi$ from $\boldsymbol{x}_0$.
It is expressed as an expectation over the Bernoulli random cost from \Cref{eq:cost}:
\begin{equation}
    r^\pi_{\boldsymbol{x}_0}(\mathcal{S},\mathcal{O}) = \mathbb{E}^\pi_{\boldsymbol{x}_0} \left[ \sum^{N-1}_{j=0} \left( \prod^{j-1}_{i=0} \boldsymbol{1}_{\mathcal{O}\setminus\mathcal{S}}(\boldsymbol{x}_i) \right) \boldsymbol{1}_{\mathcal{X}\setminus\mathcal{O}}(\boldsymbol{x}_j) \right],
    \label{eq:risk}
\end{equation}
where $\mathbb{E}^\pi_{\boldsymbol{x}_0} \left[\cdot\right]$ is the expectation operator of the (random) cost associated with an agent starting at $\boldsymbol{x}_0$ and following policy $\pi$.
In this paper, we pay a special attention to how such policies are found in practice using dynamic programming over a discretized state space, which leads to risk approximation errors.

\section{Dynamic Programming for Conservative Risk Inference}
\label{sec:planning_approach}

We now define a general infinite horizon dynamic programming (DP)-based solution to our reachability problem over the hybrid state space.
Then, we propose a numerical algorithm that, under certain conditions, 
conservatively accounts for risk prediction errors when finding maximally safe policies.

\subsection{General Infinite Horizon DP Solution}

Given an arbitrary policy $\pi$, the risk function defined by \Cref{eq:risk} can be computed using a backward recursion rule.
This recursion admits a DP algorithm to find a policy that maximizes the probability of safety from any initial state, as detailed in~\cite[Theorem 6]{summers_verification_2010}.
We optimize over an infinite horizon, that is, until value function convergence, and justify this decision based on the following observations:
\begin{enumerate}
    \item Our hybrid state space $\mathcal{X}$ has a temporal dimension and every available action has a positive duration.
    \item The operational region $\mathcal{O}$ (and consequently, the safe set $\mathcal{S}$) is bounded.
\end{enumerate} 
Therefore, the rover always reaches a terminal state (entering $\mathcal{S}$ or exiting $\mathcal{O}$) in finite time.

For our problem, a value iteration-based algorithm begins by initializing the value function with
\begin{equation}
    {V_N(\boldsymbol{x}) = \boldsymbol{1}_{\mathcal{X}\setminus\mathcal{S}}(\boldsymbol{x}) \quad \forall \boldsymbol{x} \in \mathcal{X}},
\end{equation}
which maps all safe states to zero and to one otherwise.
Then, the following backward recursion is run until the value function converges:
\begin{equation}
\begin{aligned}
    V_{k}(\boldsymbol{x}) =
     \boldsymbol{1}_{\mathcal{X}\setminus\mathcal{O}}(\boldsymbol{x}) + \boldsymbol{1}_{\mathcal{O}\setminus\mathcal{S}}(\boldsymbol{x})\min_{a \in \mathcal{A}%
     }\mathbb{E}\left[V_{k+1}(\boldsymbol{f}(\boldsymbol{x},a))\right]\quad  \forall \boldsymbol{x} \in \mathcal{X}.
\end{aligned}
\end{equation}

Once value iteration has converged, the (optimal) value function represents the lowest probability of failure (i.e., the probability of exiting the operational region before reaching the safe set) from any state:
\begin{equation}
    r_{\boldsymbol{x}}^{\pi^*}(\mathcal{S}, \mathcal{O}) = V(\boldsymbol{x}) \quad \forall \boldsymbol{x} \in \mathcal{X}.
\end{equation}
A corresponding (optimal) policy $\pi^*$ for all non-terminal states is obtained as
\begin{equation}
    \pi^*(\boldsymbol{x}) =
     \argmin_{a \in \mathcal{A}
     }
     \mathbb{E}\left[V(\boldsymbol{f}(\boldsymbol{x},a))\right] \quad \forall \boldsymbol{x} \in \mathcal{O}\setminus\mathcal{S}.
\end{equation}

\subsection{State Space Discretization and Mapping}
Following a similar approach to the numerical implementations of DTSHS reachability  reviewed in \Cref{sec:relatedwork}, we discretize the continuous components of the state space to make the (approximate) policy calculation tractable.
We divide the time-energy domain of the operational region of the state space (${\mathcal{T} \times \mathcal{B} \subset\mathbb{R}_{\geq\,0} \times \mathbb{R}_{\geq\,0}}$) into non-overlapping partitions by discretizing the temporal and energy axes into $n_\mathcal{T}$ and $n_\mathcal{B}$ bins, respectively.
Moreover, we represent all failure states in $\mathcal{X}\setminus\mathcal{O}$ as a single terminal sink.
The resulting, complete discretized state space $\mathcal{Z}$ has cardinality ${    |\,\mathcal{Z}\,| = |\,\mathcal{C}\,|\times n_\mathcal{T} \times n_\mathcal{B} + 1}$.

To produce an approximate policy for the underlying hybrid system, a deterministic mapping from any hybrid state to the discretized state space is required.
Let ${\boldsymbol{\phi}: \mathcal{X} \rightarrow \mathcal{Z}}$ be such a (surjective) function.
This map plays a central role in how approximation errors are propagated during the calculation of the (approximately optimal) value function.
For instance, in~\cite{abate_computational_2007}, a paradigm akin to nearest neighbour selection is used.
Given this mapping and a discretization resolution, Lipschitz continuity assumptions on state transition dynamics allow for the calculation of a bound on the risk prediction error.
Instead of calculating error bounds for our specific problem instance, in this paper we propose a map that, under certain conditions, accounts for approximation errors in a conservative manner.

We inspect the risk behaviour along each continuous dimension of our state space.
The derivative of risk with respect to a rover's energy level is always negative or zero.
Assuming all other state values are held constant, storing more energy will never increase the probability of failing to reach the safe region.
Therefore, a conservative function $\boldsymbol{\phi}$ should always map a hybrid state to an energy bin with a value equal or less than that of the hybrid state.

\begin{figure}[t]
    \centering
    \includegraphics[width=0.9\columnwidth]{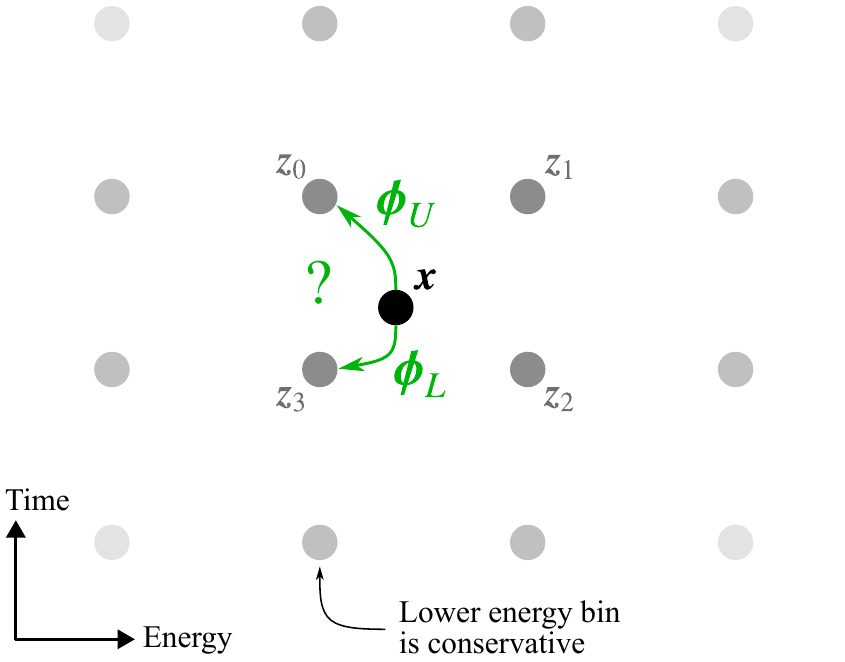}
    \caption{Projection of the discretized state space $\mathcal{Z}$ and a hybrid state $\boldsymbol{x}$ on the time-energy plane.
    The four neighbouring discretized states are labelled as $\boldsymbol{z}_{0...3}$ and each is associated with a different time-energy partition.
    A map that seeks to overestimate the value function (risk, in this case) from a hybrid state should always map to the lowest energy bin.
    Depending on where the hybrid state is in the state space, a conservative map might use either lower time bin ($\boldsymbol{z}_3$, with $\boldsymbol{\phi}_L$) or the upper one ($\boldsymbol{z}_0$, with $\boldsymbol{\phi}_U$).} 
    \label{fig:grid}
\end{figure}

The relationship between risk and the temporal dimension is not as straightforward, however: risk can vary in a nonmonotonic fashion with time.
This is due to the time-varying insolation conditions across the lunar surface (shown for one grid cell in \Cref{fig:safehaven}).
As such, the temporal bin to which a conservative function $\boldsymbol{\phi}$ should assign a hybrid state depends on where exactly this state is located in the state space.
It is difficult to define a single, conservative mapping function ahead of time (i.e., before solving for the approximate, maximally-safe policy) since it is unclear where in the state space the risk derivative with respect to time will be positive and negative.
This temporal mapping ambiguity is illustrated in \Cref{fig:grid}.
We define two functions: $\boldsymbol{\phi}_L$, which maps a hybrid state $\boldsymbol{x}$ to the lower temporal bin, and $\boldsymbol{\phi}_U$, which maps $\boldsymbol{x}$ to the upper temporal bin.
Both functions map to the lower energy bin which, as explained above, is always a conservative approximation.

\subsection{Conservative Approximate DP Solution}

Due to the ambiguity that arises by having two candidate mapping functions $\boldsymbol{\phi}_L$ and $\boldsymbol{\phi}_U$, we suggest accounting for both of them simultaneously during the optimization process.
We propose a min-max value iteration methodology to conservatively account for value function approximation errors.
Our approach begins by initializing the approximate value function as
\begin{equation}
    \hat{V}_N(\boldsymbol{z}) = \boldsymbol{1}_{\mathcal{X}\setminus\mathcal{S}}(\boldsymbol{z}) \quad \forall \boldsymbol{z} \in \mathcal{Z}.
\end{equation}
Then, the following recursion takes place:
\begin{equation} \label{eq:recursion}
\begin{aligned}
     \hat{V}_{k}(\boldsymbol{z}) =
     \boldsymbol{1}_{\mathcal{X}\setminus\mathcal{O}}(\boldsymbol{z}) + \boldsymbol{1}_{\mathcal{O}\setminus\mathcal{S}}(\boldsymbol{z})\min_{a \in \mathcal{A}
     } \max_{\boldsymbol{\phi} \in \Phi} \mathbb{E}\left[\hat{V}_{k+1}(\boldsymbol{\phi}(\boldsymbol{f}(\boldsymbol{z},a))) \right] \\ \forall \boldsymbol{z} \in \mathcal{Z},
\end{aligned}
\end{equation}
where $\Phi = \{\boldsymbol{\phi}_L, \boldsymbol{\phi}_U\}$.%
As such, every iteration chooses the best action (associated with the lowest risk) assuming the worst map from $\Phi$ applies.
This recursion continues until the following convergence criterion is met,
\begin{equation}
    \max_{\boldsymbol{z} \in \mathcal{Z}} | \hat{V}_{k+1}(\boldsymbol{z}) - \hat{V}_{k}(\boldsymbol{z})| \leq \varepsilon,
\end{equation}
where $\varepsilon$ is a (usually very small) user-defined threshold.

The conservativeness of the approximate value function is dependent on the discretization resolution and the true risk function $V$.
Precisely, $\hat{V}(\boldsymbol{z})$, the approximate value function computed for a discrete state $\boldsymbol{z} \in \mathcal{Z}$, must upper-bound the true risk at any hybrid state that is mapped to that discrete state.
The discretization along the temporal dimension is the most important since risk can vary nonmonotonically with time.
We empirically validate conservativeness with reasonable discretization resolutions in our experiments (see \Cref{sec:experiments}.
A systematic selection of the state space discretization resolution for specific environment models is kept as future work. 

We point out that our optimization framework is compatible with other game-theoretic approaches in the literature.
For instance, the approach in~\cite{ding_stochastic_2013} specifically accounts for an adversarial agent, while~\cite{yang_dynamic_2018} makes use of an ambiguity set over possible disturbance distributions.
From a practical stand point, all of these approaches modify the state transition probabilities, just like our approach.

\section{Experiments}
\label{sec:experiments}

We demonstrate our method by conducting stochastic reachability analyses in areas containing PSRs near Cabeus Crater at lunar south pole.
We first provide details about our experimental setup and then discuss the results obtained.\footnote{Additional experimental results are available at the following URL:\linebreak \url{https://papers.starslab.ca/recovery-policies-psr-exploration}.}

\subsection{Experimental Setup}

We employ an hourly solar visibility dataset generated for the Cabeus region at the lunar south pole, provided by the NASA Jet Propulsion Laboratory.
The dataset contains georeferenced maps with a resolution of 240 metres per pixel that coarsely indicate the instantaneous percentage of the solar disk that is visible (0\%, 20\%, \dots, 100\%).
Visibility calculations account for the Moon's curvature and possible occlusions caused by obstacles within the Cabeus area and beyond.
The time window covered by the dataset stretches from August 1 to October 27, 2029 (three lunar synodic days, 2,089 maps in total).\footnote{All date and time references are with respect to the Coordinated Universal Time (UTC). The corresponding timestamps expressed as seconds past the UNIX epoch (i.e., UNIX time) are sometimes specified as the experiments rely on this representation.}
These maps express the solar visibility as seen from a height of two metres above the surface (the height of the solar panels of our hypothetical rover) and we derive the corresponding irradiance maps assuming a constant total solar flux of 1,367 W/m$^\text{2}$.
This value is the solar constant at a distance of 1 AU from the Sun and it is commonly employed in conceptual mission studies for the lunar surface~\cite{della_torre_amalia_2010, stoica_transformers_2017}.

\begin{figure}[t]
    \centering
    \includegraphics[width=\columnwidth]{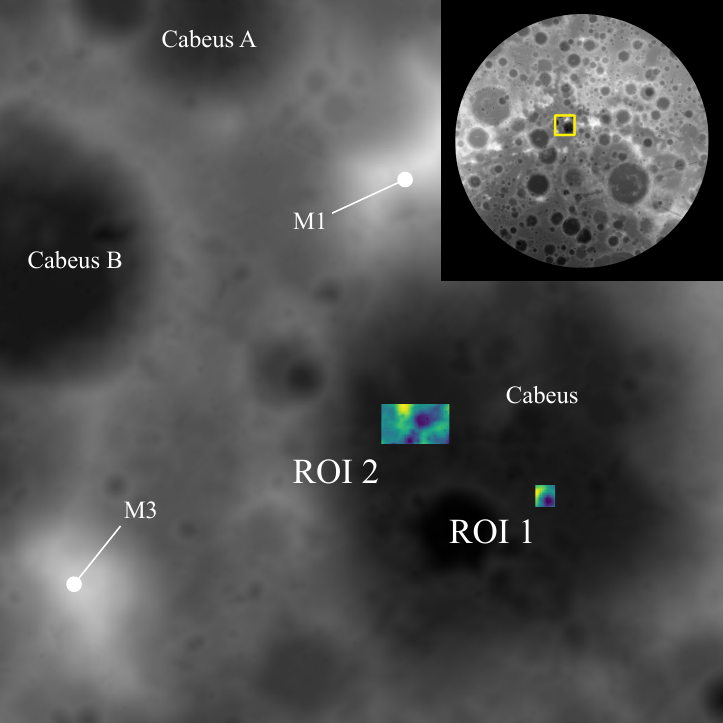}
    \caption{Elevation model of the Cabeus region spanning a 143 kilometres-wide, square area. Named mountains and craters are identified for reference. The global context of this map is shown at the top right by the stereographic projection of the south pole, stretching to a latitude of 60 degrees south. The specific regions of interest (ROI) chosen for our experiments are colourized.}
    \label{fig:cabeus_maps}
\end{figure}

For terrain assessment, we use a high-resolution elevation model of the lunar south pole retrieved from NASA's Moon Trek database~\cite{day_moon_2017}.
This model is composed of several high-resolution (20 metres per pixel) data products from the Chang'e 2 spacecraft~\cite{li_lunar_2018} that have been mosaicked together, projected to a stereographic polar format, and registered horizontally and vertically against NASA's Lunar Orbiter Laser Altimeter data.
The elevation data is cropped to the Cabeus region only and its resolution is reduced to 240 metres per pixel to match that of the insolation data described above. 
The corresponding slope and aspect maps are generated using Horn's method as implemented in the Geospatial Data Abstraction Library~\cite{gdalogr_contributors_gdalogr_2022}.
Navigation is allowed only to grid cells with a slope magnitude below 20 degrees.
\Cref{fig:cabeus_maps} shows the global context of the Cabeus area and the specific regions of interest (ROIs) containing the data used in our experiments.
Our first and second experiments (\Cref{sec:exp1,sec:exp2}) use data from ROI 1, while our third experiment (a traverse from the LCROSS impact area, \Cref{sec:exp3}) uses data from ROI 2.

We assume that the rover maintains a constant area of its solar panels oriented towards the Sun regardless of the rover's heading.
As explained previously, this approximately represents a rover with tilted panels evenly distributed on all its sides or on an actuated platform.
We leave improvements to this model, such as accounting for a varying rover attitude over different terrain geometries or the use of an asymmetric panel layout, as future work.
When the rover is in motion, we assume that it travels at a constant velocity and that the mobility power consumption is constant, independent of the rover's roll and pitch.
Additionally, a constant background power draw is required for housekeeping tasks.
The rover can hibernate in place (requiring a lower background power load) upon successfully reaching a safe state.

Drive time and energy consumption between neighbouring grid cells depends solely on the physical distance between their centres, which in turn depends on their relative position and elevation difference.
In addition to driving to one of the eight neighbouring cells, the rover can also wait in place for a fixed duration of 5,000 seconds.
We provide our own Python library called \texttt{gplanetary\_nav}, which we use for queries involving terrain and insolation maps, as open source software.\footnote{Queries to \texttt{gplanetary\_nav} include estimates of driving distances, duration and energy drawn, and the instantaneous solar power generated from specific spatiotemporal states. The library repository is available here: \\ \url{https://papers.starslab.ca/recovery-policies-psr-exploration}}
All of our experiments are carried out on a computer with an AMD Threadripper 2920X 3.5 GHz 12-Core CPU and 128 GB of memory running Ubuntu 18.04.

\subsection{Baseline Policies}

\begin{figure}[t]
    \centering
    \includegraphics[width=0.9\columnwidth]{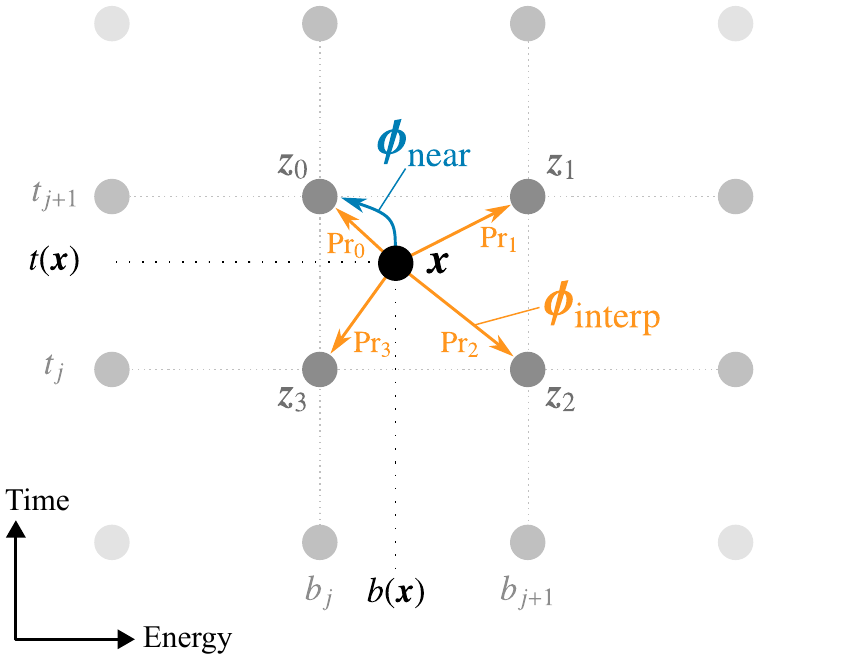}
    \caption{Behavior of the ``nearest'' and ``interpolation'' maps from the hybrid space $\mathcal{X}$ to the discretized space $\mathcal{Z}$ employed to compute the policies against which we compare our approach. The probabilities associated with the ``interpolation'' map are indicated with $\text{Pr}_\#$ (short for $\text{Pr}(\boldsymbol{z}_\#\,|\,\boldsymbol{x})$, which is defined in \Cref{eq:probabilities}).} 
    \label{fig:other_policies}
\end{figure}

We compare our conservative approximate dynamic programming method against two other DP approaches that use different state space mappings.
The behaviour of the alternative approaches is illustrated in \Cref{fig:other_policies}.

First, what we refer to as the ``nearest'' policy is calculated using a deterministic map from any (hybrid) state $\boldsymbol{x}$ to the nearest grid state ${\boldsymbol{z} = \boldsymbol{\phi}_{\text{near}}(\boldsymbol{x})}$, which is similar to the method in~\cite{abate_computational_2007}.
In the current implementation, $\boldsymbol{z}$ is the discrete state sharing the same  grid cell location as $\boldsymbol{x}$ and with the closest time and energy values.

Second, the ``interpolation'' policy relies on a multi-linear interpolation scheme to relate any hybrid state $\boldsymbol{x}$ to nearby grid states, akin to the experimental setup in~\cite{chapman_risk-sensitive_2019}.
Given a (hybrid) state $\boldsymbol{x}$, the four nearest grid states are those sharing the same grid cell on the orbital map and defining a hull that encloses $\boldsymbol{x}$ in the time-energy plane.
Let $t_j$ and $t_{j+1}$ be the two nearest discrete time values such that ${t_j \leq t(\boldsymbol{x}) < t_{j+1}}$, and let $b_j$ and $b_{j+1}$ be the two nearest discrete energy values such that ${b_j \leq b(\boldsymbol{x}) < b_{j+1}}$.
Let $\Omega_{\boldsymbol{x}} = \{ \boldsymbol{z}_{0...3}\}$ be the set of four discrete states near $\boldsymbol{x}$ (as shown in \Cref{fig:other_policies}).
We define the corresponding mapping ${\boldsymbol{\phi}_{\text{interp}}(\boldsymbol{x})}$ on a probability space.
The probability ${\text{Pr}(\boldsymbol{z}\,|\,\boldsymbol{x}) \; \forall \boldsymbol{z} \in \Omega_{\boldsymbol{x}}}$ is calculated using the normalized distances on the time-energy plane between the hybrid state $\boldsymbol{x}$ and the corresponding neighbouring grid states $\boldsymbol{z}$,
\begin{align}
\begin{split}
  \text{Pr}(\boldsymbol{z}_0\,|\,\boldsymbol{x}) &= \left( 1-\frac{b(\boldsymbol{x})-b_j}{b_{j+1}-b_j} \right) \left( \frac{t(\boldsymbol{x})-t_j}{t_{j+1}-t_j} \right), \\
  \text{Pr}(\boldsymbol{z}_1\,|\,\boldsymbol{x}) &= \left( \frac{b(\boldsymbol{x})-b_j}{b_{j+1}-b_j} \right) \left( \frac{t(\boldsymbol{x})-t_j}{t_{j+1}-t_j} \right), \\
  \text{Pr}(\boldsymbol{z}_2\,|\,\boldsymbol{x}) &= \left( \frac{b(\boldsymbol{x})-b_j}{b_{j+1}-b_j} \right) \left( 1-\frac{t(\boldsymbol{x})-t_j}{t_{j+1}-t_j} \right), \\
  \text{Pr}(\boldsymbol{z}_3\,|\,\boldsymbol{x}) &= \left( 1-\frac{b(\boldsymbol{x})-b_j}{b_{j+1}-b_j} \right) \left( 1-\frac{t(\boldsymbol{x})-t_j}{t_{j+1}-t_j} \right).
\end{split}
\label{eq:probabilities}
\end{align}
Since our action space is discrete (meaning that we cannot combine the action of each neighbouring grid state using a weighted sum), the optimal policy obtained with the interpolation map is stochastic.

\subsection{Comparison of Approximate Methods}
\label{sec:exp1}

\begin{figure}
    \centering
    \includegraphics[width=\columnwidth]{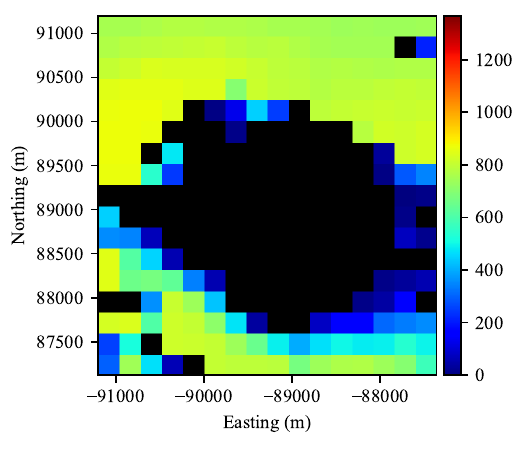}
    \caption{The average irradiance map (in W/m\textsuperscript{2}, pixel resolution of 240 metres) of ROI 1 (shown in \Cref{fig:cabeus_maps}). Locations on the lunar surface that do not receive any sunlight over the operational time interval, indicated by black pixels, are assumed to be PSRs.}
\label{fig:exp1_map}
\end{figure}

Our first experiment is designed to empirically assess the performance of the three policies obtained with the \emph{nearest}, \emph{interpolation}, and \emph{conservative} state space mappings.
We focus on scenarios where a rover exits from a PSR to reach a safe state.
The risk predicted by each policy from different start states is compared against the actual risk, which we estimate through Monte Carlo simulations.
The experiment uses data from ROI 1 (shown in \Cref{fig:cabeus_maps}).
The average solar irradiance map for ROI 1 over the designated operational time interval is shown in \Cref{fig:exp1_map}.
Experiment parameters, such as the rover model, state space boundaries, noise model, and approximate dynamic programming parameters, are listed in \Cref{tab:exp1_params}.

\begin{table*}[t]
\centering

\begin{tblr}{cells={l}, row{1} = {rowsep=3pt}} %
\toprule
\textbf{Parameter}                      & \textbf{Value}                                        \\ \midrule

\TCategoryAboveSpace\textbf{Operational state space} &                                                       \\
Operational time interval               & 22 Aug 2029 04:33:20 to 26 Aug 2029 04:33:00 (4 Earth days) \\
Operational energy interval             & 500 Wh to 10,000 Wh                               \\

\TCategoryAboveSpace\textbf{Safe state space subset}        &                                                       \\
Safe haven locations                   & Any non-PSR orbital grid cell is a candidate                       \\
Time limit                              & Same as operational time limit everywhere             \\
Minimum energy required by time limit   & 1,000 Wh everywhere                                 \\

\TCategoryAboveSpace\textbf{Approximate DP details}         &                                                       \\
Time discretization resolution          & 3,600 seconds (1 hour)                                \\
Energy discretization resolution        & 100 Wh                                              \\
Total number of discretized states      & Approximately 2.5 million                             \\
Convergence criterion ($\varepsilon$)                   & 1e-5 \\
\TCategoryAboveSpace\textbf{Random faults / noise model}    &                                                       \\
Average fault spatial rate              & 1 every 1,000 m driven                                \\
Fault recovery duration                 & 18,000 seconds (5 hours)                              \\

\TCategoryAboveSpace\textbf{Rover model}                    &                                                       \\
Solar panel area                        & 1.5 m\textsuperscript{2}             \\
Solar panel efficiency                  & 30 \%                                                 \\
Driving velocity                        & 0.10 m/s                                              \\
Driving power draw                     & 220 W                                                 \\
Fault resolving power draw                        & 80 W    \\
Idling power draw (waiting in place)               & 80 W                                                  \\
Idling power draw (hibernating)          & 40 W                                                  \\
Battery capacity                        & 10,000 Wh    \\ [2mm] \bottomrule                                      
\end{tblr}
\caption{Parameters for the first and second experiments (\Cref{sec:exp1,sec:exp2}, respectively).}
\label{tab:exp1_params}
\end{table*}

A total of 100 simulation batches are run, each according to the following procedure:
\begin{enumerate}
    \item Randomly sample a start state inside a PSR and within the operational subset of the state space.
    \item Generate a random spatial fault profile by sampling from the underlying Poisson process corresponding to the disturbance model.
    \item Roll out all three policies against this fault profile from the start state.
    \item Repeat (from Step 2 onward) one million times from the same start state.
\end{enumerate}
To avoid trivial cases (start states that are overwhelmingly safe or from which the safe region is definitely out of reach), we only retain start states with a predicted risk between 1\% and 90\% according to all three policies.
A trial leaving the operational region of the state space or entering a state from which the predicted risk is 100\% is labelled as a failure.
A trial that reaches the safe region first is labelled as a success.
For every simulation batch, the actual risk associated with a start state and a policy is, approximately, the ratio of failed trials.

To avoid very small discrepancies caused by sampling, a risk prediction that is less than the actual risk by more than 0.1\% is deemed reckless (i.e., dangerous).
This margin is an order of magnitude lower than the scale of risk mispredictions we consider significant.

\begin{figure}[t]
    \centering
    \includegraphics[width=\columnwidth]{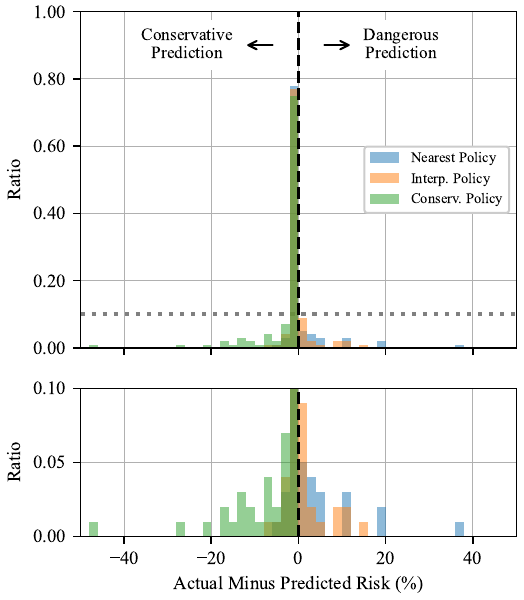}
    \caption{Overview of results from Experiment 1. \textit{Top}: Histogram of the actual risk estimate minus the predicted risk for 100 simulation batches, grouped into 2 percent-wide bins. For each policy, the histogram bars are normalized such that their cumulative height sums to 1.0. \textit{Bottom}: a close up view of the area of the top plot below the horizontal dotted gray line.}
    \label{fig:exp1_histplot}
\end{figure}

Histograms of the difference between the actual risk and the predicted risk for all simulation batches and for each policy are shown in ~\Cref{fig:exp1_histplot}.
A negative difference is indicative of a conservative risk prediction while a positive difference corresponds to a dangerous (underconservative or optimistic) one.
A difference close to zero implies an accurate prediction relative to the true risk.
As illustrated in \Cref{fig:exp1_histplot}, all dangerous predictions are made by the policies using the nearest and interpolation maps.
Predictions made with the policy computed using our proposed approach are conservative.
The degree of conservatism is influenced by the discretization resolution (which affects the magnitude of approximations at each planning step).
Another contributing factor is the number of steps/decisions taken to reach safety; approximation errors compound with trial length.
We further observe the influence of discretization resolution in \Cref{sec:exp2} and trial length in \Cref{sec:exp3}.

\begin{figure}[t]
    \centering
    \includegraphics[width=\columnwidth]{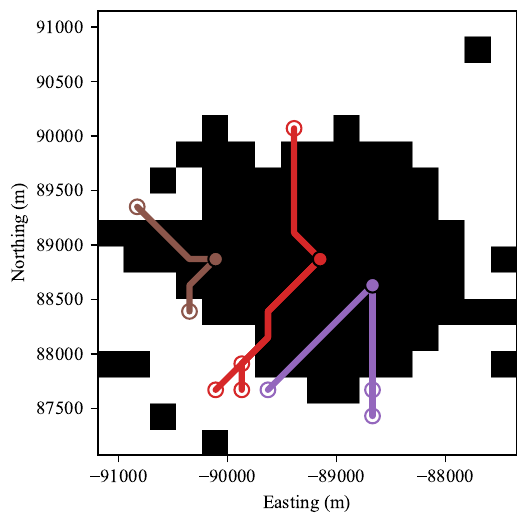}
    \caption{Successful trials from different start states sharing the same location on the orbital map (indicated with filled circles). The end of each path is marked by a hollow circle. The background is a binary mask of the average solar irradiance data shown in \Cref{fig:exp1_map}. All results were produced by our conservative policy.}
    \label{fig:exp1_spatiotemporal_paths}
\end{figure}

\begin{figure}[t]
    \centering
    \includegraphics[width=\columnwidth]{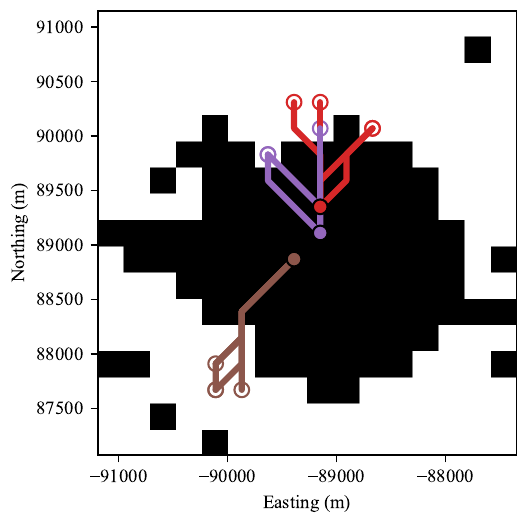}
    \caption{Successful trials originating from the same start state (i.e., the same location, departure time, and battery energy), indicated with filled circles. Each trial involves a different fault profile. The left-most and right-most end locations (hollow circles) of the trials shown in red are almost 1 kilometre apart. The background is a binary mask of the average solar irradiance data shown in \Cref{fig:exp1_map}.} 
    \label{fig:exp1_stochastic_paths}
\end{figure}

The importance of spatiotemporal planning is illustrated in \Cref{fig:exp1_spatiotemporal_paths}.
Successful trials originating from the same location (but with different start times and energies) might have drastically different recovery strategies due to the time-varying insolation conditions outside of the PSR.
Furthermore, the safest recovery strategy also varies based on \textit{when} and \textit{where} faults occur during the traverse.
Policy-based (online) planners, which proactively account for delays and inherently enable the rover to adapt its behaviour on the fly, provide a clear safety benefit over conventional offline spatiotemporal planners.
The diversity of successful paths from identical start states is shown in \Cref{fig:exp1_stochastic_paths}.
Despite being relatively short (on the order of a kilometre), these simulated drives end in locations that can be almost a kilometre apart, a significant distance for a rover with an effective traverse speed of 0.10 m/s.

\begin{figure}[!b]
    \centering
    \includegraphics[width=\columnwidth]{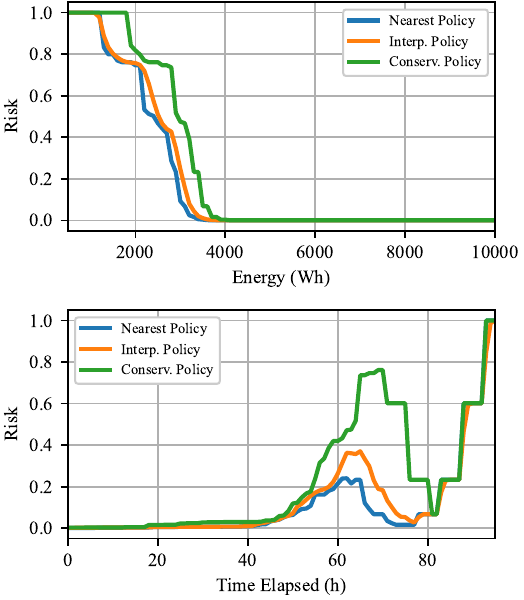}
    \caption{Variation of the predicted risk along the continuous dimensions of the state space according to all three policies. The discretization resolution of each dimension is indicated in \Cref{tab:exp1_params}. \textit{Top}: risk as a function of energy from grid cell (9,7) at 21:23:33 on August 24 2029 (UNIX timestamp 1882301013 seconds). \textit{Bottom}: risk as a function of time (relative to August 22 2029 at 04:33:20, or UNIX timestamp 1882067600 seconds) from grid cell (9,7) and a constant energy value of 2,875 Wh.} \label{fig:exp1_risk_variation}
\end{figure}

We also illustrate how the value functions (i.e., the risk functions) corresponding to all three policies vary along the continuous dimensions of the state space.
For a fixed grid cell and timestamp, we plot the predicted risk as a function of energy. 
Likewise, we plot the predicted risk as a function of time given a fixed grid cell and energy.
As depicted in \Cref{fig:exp1_risk_variation}, the predicted risk varies monotonically with energy and nonmonotonically with time.
This visualization provides an insight into the motivation behind our proposed approach. 
Having more energy cannot decrease the rover's likelihood of survival.
On the other hand, the safe region of the state space has a temporal dependence, causing risk values to rise and fall with respect to time.
Additionally, in \Cref{fig:exp1_risk_variation}, the predicted risk versus time profile increases (to one) towards the right of the plot, in this case because the rover cannot reach the safe region before the time limit.
Both plots also illustrate that our proposed policy 
is conservative by design: our predicted risk is always equal to or greater than that of the other policies.

Lastly, the bottom plot in \Cref{fig:exp1_risk_variation} illustrates why the conservative behaviour 
of our approach is conditional on assumptions about the underlying risk function and the time discretization resolution.
A very coarse discretization may not properly capture the local risk maximum in the second half of the plot (near the 60--70 hours mark) and lead to dangerous risk predictions for this region of the state space.
In practice, determining whether the temporal discretization resolution is fine enough to satisfy this condition is problem-dependent and nontrivial to determine; we leave this challenge as future work.

\begin{figure*}[p]
    \centering
    \includegraphics[width=\textwidth]{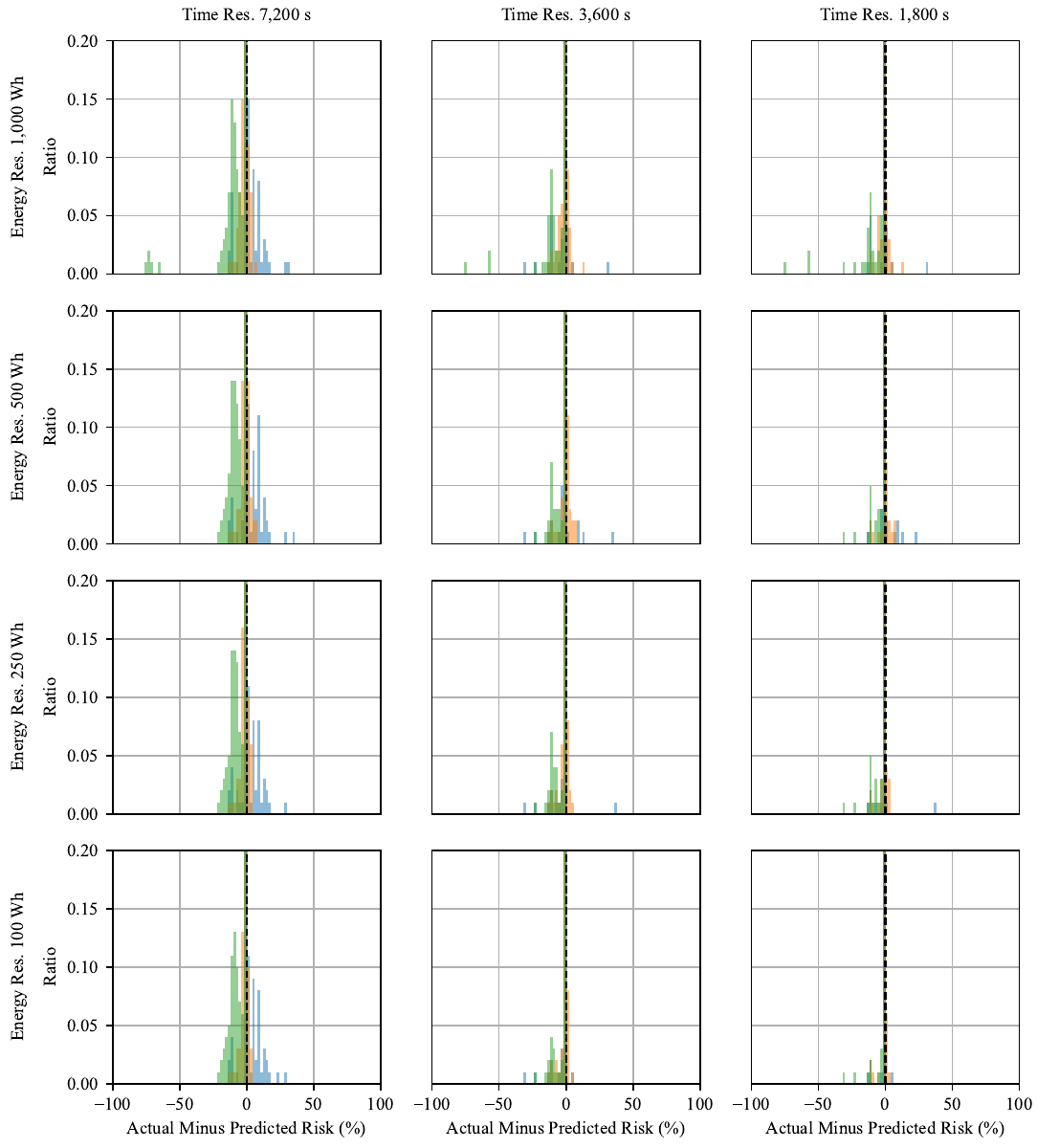}
    \caption{Histograms of the actual risk estimate minus the predicted risk for 100 simulation batches, grouped into 2 percent-wide bins. Every column shares the same discretization resolution along the temporal dimension (increasing from left to right) while every row shares the same discretization resolution along the energy dimension (increasing from top to bottom). Bar colours indicate different policies (nearest is blue, interpolation is orange, conservative is green). For each policy, the bar heights are normalized to sum to 1.0 (i.e., a single bar represents the ratio of samples falling in the corresponding histogram bin). Note that every subfigure is cropped vertically to focus on relevant results.}
    \label{fig:exp2_histplots}
\end{figure*}

\subsection{Effect of Discretization Resolution}
\label{sec:exp2}

The second experiment consists in visualizing the effect of different discretization resolutions on the riskiness of predictions of the nearest, interpolation and conservative policies.
This experiment uses data from the same region of Cabeus Crater (ROI 1).
The experiment employs the same parameters and models (shown in \Cref{tab:exp1_params}) except for the time and energy discretization resolutions, which are now variables.
We conduct this experiment using time discretization resolutions of 7,200, 3,600 and 1,800 seconds (corresponding to 2, 1 and 0.5 hours, respectively) and energy discretization resolutions of 1,000, 500, 250 and 100 Wh.
This results in 12 possible combinations of time and energy discretization resolutions.

The experiment begins by generating one million random spatial fault profiles by sampling the Poisson process corresponding to the disturbance model parameters in \Cref{tab:exp1_params}.
Then, a total of 100 simulation batches are carried out, each according to the following procedure:
\begin{enumerate}
    \item Randomly sample a start state inside a PSR and within the operational subset of the state space.
    \item Roll out all three policies for a specific combination of time and energy discretization against each (pregenerated) fault profile from the current start state.
    \item Repeat the previous step for all the other time and energy discretization combinations.
\end{enumerate}

As with the previous experiment, we only retain start states with a predicted risk between 1\% and 90\% according to all three policies for all possible time and energy discretization combinations.
The success of individual trials and the actual risk associated with every simulation batch are calculated the same way as described in \Cref{sec:exp1}.
We highlight that the random start states and fault profiles are purposefully the same for each time and energy discretization combination.
Variations in the results within the same batch are thus only caused by different discretization resolutions and isolated from other factors.

The histograms of risk differences for each discretization setting are shown in \Cref{fig:exp2_histplots}.
Globally, there is a clear trend: increasing the discretization resolution (along both the time and energy dimensions) causes risk differences to converge towards 0 for all policies and drives down conservatism.
This observation is in accordance with previous theoretical results on the topic~\cite[Section 5]{abate_computational_2007}.
Assuming that the temporal discretization resolution is sufficiently fine, our proposed policy converges in a conservative manner (i.e., the risk difference is always negative or zero) while this is not the case for the other policies.

At a local scale, smaller increments in the time and/or energy discretization resolution do not always reduce the degree of conservatism for every possible state.
Each discretization combination is associated with a different grid layout in the continuous state space and, consequently, a different map from the hybrid to the discretized space.
The distance in continuous space between a hybrid state and its corresponding discrete state may sometimes increase (and therefore increase the approximation magnitude) despite an increasing discretization resolution.

Notably, we observe that the conservatism (or riskiness) caused by a relatively coarse time or energy discretization somewhat outweighs the effects of a finer discretization along the other axis.
In \Cref{fig:exp2_histplots}, the histograms in the first column (corresponding to a  coarse temporal grid resolution) remain relatively unchanged with an increasing discretization resolution along the energy dimension.
The effect of a varying energy bin resolution is visible in the last column, which corresponds to a finer temporal discretization resolution.
Similarly, the histograms in the top row (corresponding to constantly coarse energy grid resolution) are less affected by an increasing time discretization resolution than those in the bottom row, which correspond to a constantly fine energy discretization resolution.

We mention that the plot with the same discretization settings as the first experiment (3,600 seconds and 100 Wh) looks different from the plot shown in \Cref{fig:exp1_histplot}.
In the current experiment, we force all start states to be associated with nontrivial solutions for all possible discretization resolutions.
While this further constrains the regions in the state space from which the simulations are run, the choice allows us to isolate the effects of different discretization resolutions.%

\begin{figure}[h]
    \centering
    \includegraphics[width=\columnwidth]{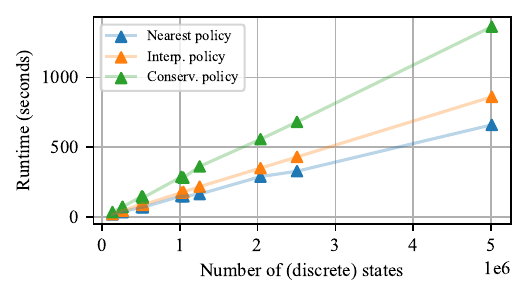}
    \caption{
        Policy generation times for different (discretized) state space sizes and state space mappings.
        The markers indicate data points retrieved from every policy generated for \Cref{sec:exp2}.
        Semi-transparent lines are added between consecutive data points for visualization purposes only.
    }
    \label{fig:exp2_runtimes}
\end{figure}

Increasing the time or energy bin resolution enlarges the corresponding, discretized state space and increases the computational effort required to  generate policies.
Here, generating a policy entails constructing the corresponding system of equations (initial value function vector and state transition probability matrices for all actions) and running value iteration until convergence.
\Cref{fig:exp2_runtimes} shows the time required to generate the policies with which data in \Cref{fig:exp2_histplots} was obtained.
Markers indicate measured time values with our implementation and lines are added between consecutive data points for visualization purposes.
In our pure Python implementation, the majority of computation time is dedicated to the construction of the state transition matrices.
This operation involves significant overhead since each matrix is constructed one row at a time.
Another contributing factor is the sparsity of the state transition matrices: those associated with the nearest map have at most three non-zero elements per row and a little more for the interpolation map.
Conservative policies require the computation of twice as many state transition matrices in order to generate the ambiguity set of mapping functions $\Phi$ introduced in \Cref{eq:recursion}.%
\section{A Practical Example: The LCROSS Impact Site}
\label{sec:exp3}

In addition to supporting online global trajectory planning, recovery policies can also be used to estimate the tradeoff in risk associated with different mission design choices.
We evaluate the safety implications of multiple rover mobility models in the context of a long drive departing from the Lunar Crater Observation and Sensing Satellite (LCROSS) impact site at the lunar south pole.
While the experiments in previous sections benchmarked our approach against existing policies in the context of small- to medium-scale recovery drives, this last experiment simulates a long-range, multi-kilometre traverse.
We consider our conservative policy and apply it in a region of Cabeus crater receiving little to no sunlight.

\subsection{Experimental Setup}

\begin{figure}[b!]
\centering
\includegraphics[width=\columnwidth]{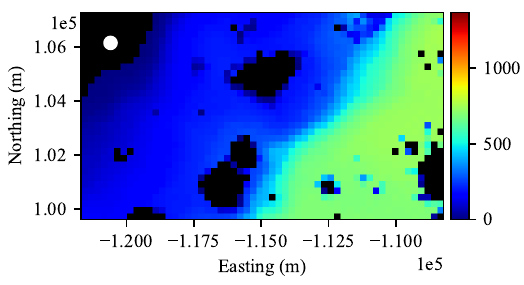}
\caption{The average irradiance map (in W/m\textsuperscript{2}, pixel resolution of 240 metres) of the region of interest for the third experiment. The figure shows a close up view of ROI 2 from \Cref{fig:cabeus_maps}, spanning an area that measures approximately 7.7 by 13.4 kilometres. Black pixels do not receive any sunlight over the operational time interval. The white dot indicates the start location for all of the simulations.}
\label{fig:exp3_map}
\end{figure}

Our experiment uses data from ROI 2 in \Cref{fig:cabeus_maps}.
The average solar irradiance map of this region over the chosen operational time window is shown in \Cref{fig:exp3_map}.
The start location of every drive (situated in a PSR) is also indicated.
We refer to this location as the `LCROSS impact site' from now on.

\begin{table*}[t]
\centering
\begin{tblr}{cells={l}, row{1} = {rowsep=3pt}} %
\toprule
\textbf{Parameter}                      & \textbf{Value}                                        \\ \midrule

\TCategoryAboveSpace\textbf{Operational state space} &                                                       \\
Operational time interval               & 23 Aug 2029 18:33:20 to 31 Aug 2029 21:33:20 (roughly 8 Earth days) \\
Operational energy interval             & 500 Wh to 30,000 Wh                               \\

\TCategoryAboveSpace\textbf{Safe state space subset}        &                                                       \\
Safe havens locations                   & Orbital grid cell with an avg. solar irradiance $\geq$500 W/m\textsuperscript{2} is a candidate                       \\
Time limit                              & 26 Sep 2029 17:33:20 (about 1 lunar day past operational limit)             \\
Minimum energy required by time limit   & Variable, energy must be above 5,000 Whr by the end of the \textit{following} lunar day                                 \\

\TCategoryAboveSpace\textbf{Approximate DP details}         &                                                       \\
Time discretization resolution          & 3,600 seconds (1 hour)                                \\
Energy discretization resolution        & 250 Wh                                              \\
Total number of discretized states      & Approximately 41.5 million                             \\
Convergence criterion ($\varepsilon$)                   & 1e-5                                                  \\

\TCategoryAboveSpace\textbf{Random faults / noise model}    &                                                       \\
Average fault spatial rate              & 1 every 5,000 m driven                                \\
Fault recovery duration                 & 36,000 seconds (10 hours)                              \\

\TCategoryAboveSpace\textbf{Rover model}                    &                                                       \\
Solar panel area                        & 1.5 m\textsuperscript{2}   \\
Solar panel efficiency                  & 30 \%                                                 \\
Driving velocity                        & 0.02, 0.03, 0.05, 0.10 m/s                            \\
Driving power draw                     & 60, 90, 150, 300 W (respectively)                                \\
Fault resolving power draw              & 50 W  \\
Idling power draw (waiting in place)               & 40 W                                                  \\
Idling power draw (hibernating)          & 30 W                                                  \\
Battery capacity                        & 30,000 Wh  \\ [2mm]
\bottomrule   
\end{tblr}
\caption{Experiment 3 parameters. The four different rover models tested share the same parameters except for the effective driving velocity and power draw.}
\label{tab:exp3_params}
\end{table*}

The set of safe states $\mathcal{S}$ is defined slightly differently than for the first and second experiments.
We consider safe havens (the projection of $\mathcal{S}$ onto the orbital map) to be locations receiving an average solar irradiance of at least 500 W/m\textsuperscript{2} over the designated operational time window.
These locations are only found on the right side of the map shown in \Cref{fig:exp3_map}.
Additionally, upon reaching a safe haven, we consider the rover to be in a safe state only if it has enough energy to hibernate in place to meet a minimum energy threshold (5,000 Whr) at the end of the \textit{following} lunar day (September 26 2029 at 17:33:20, or UNIX timestamp of 1,885,138,400 seconds).
This requires the rover to be able to survive the entire lunar night, as well as being in a location that receives sunlight the following lunar day.
\Cref{tab:exp3_params} details the parameters that define the safe and operational regions of the state space, the continuous space discretization resolution, and the disturbance model.
The discretized state space is an order of magnitude larger than that of the previous experiments, with 41.5 million states.

We compare four rover models, each with a different effective drive velocity: 0.02 m/s, 0.03 m/s, 0.05 m/s and 0.10 m/s.
The power consumption for drive actions is scaled such that the driving energy consumed over a fixed travel distance is the same for all models.
We limit our study to a maximum speed of 0.10 m/s, corresponding to drive actions that last as little as 2,400 seconds given an orbital grid resolution of 240 metres.
At faster speeds, the duration of drive actions would be significantly less than the (discretized) temporal resolution of the current experiment.
The rover models are detailed at the bottom of \Cref{tab:exp3_params}.

We compute a total of four conservative recovery policies (one for each rover model) according to the parameters listed in \Cref{tab:exp3_params}.
All policies are computed using the same parameters (i.e., the same state space discretization, the same fault model, etc.) except for the rover velocity and driving power consumption.
In the remainder of this section, we study the value functions (i.e., risk predictions) of each policy from different start states (all co-located at the LCROSS impact site).
We visualize a variety of behaviours that result from different fault scenarios.

\subsection{Risk Prediction Analysis}

\begin{figure*}[t]
    \centering
    \includegraphics[width=\textwidth]{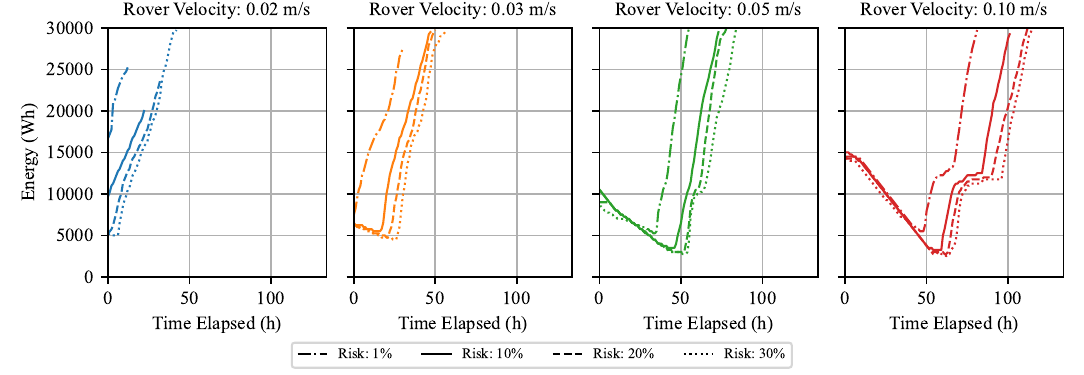}
    \caption{Minimum departure energy from the LCROSS impact site as a function of time (hours elapsed since August 23 2029 at 22:33:20, or UNIX timestamp 1,882,218,800 seconds) for different predicted risk thresholds (i.e., the probability that the rover does not reach a safe state). Each subfigure corresponds to a different policy, employing a different rover velocity model.} %
    \label{fig:exp3_minenergy}
\end{figure*}

We begin by finding the minimum energy level required for a rover departing from the LCROSS impact site as a function of the departure time such that the risk (i.e., the probability that the rover does not reach the safe region) is below a given threshold.
The plots for risk thresholds of 1\%, 10\%, 20\% and 30\% are shown in \Cref{fig:exp3_minenergy}.
Notably, increasing the risk threshold never increases the minimum energy required for a given departure time, because a lower departure energy decreases the resilience to delays.
Also, while curves corresponding to consecutive risk thresholds do not cross, they may overlap.
This simply indicates that the predicted risk increases sharply in the corresponding region of the state space and that an even more permissive risk threshold is necessary to allow for a lower departure energy.
Comparing the four subplots, the energy curves shift to the right with increasing rover speed.
This shift indicates that later departure times are permissible, since a higher effective traverse speed shortens the drive time required to reach a safe haven.

Counter to intuition, there is an increasing minimum energy requirement for early departures with velocities of 0.05 and 0.10 m/s.
We determined that this is an unintended consequence of the constant-velocity mobility models.
In this case, the entire ROI is in the shade at the beginning of the operational time interval.
Sunlight enters the ROI from the southwestern corner, quickly sweeps across the left side of the region, and then more gradually moves to the right side of the map before disappearing at the northeastern corner.
In this situation, an early departure from the LCROSS impact site (with the rover driving relatively quickly) leads to \textit{longer} traverses.
Here, \textit{longer} refers to the number of steps/actions taken to reach safety, rather than the physical distance driven.
For instance, a fast-moving rover could head eastward (forgoing some, if not all, of the opportunities for charging along the way) and then choose several `wait in place' actions upon reaching a safe haven.
Sunlight would then ``catch up'' with the rover and charge its batteries to a safe level.
\begin{figure}[!h]
    \centering
    \includegraphics[width=\columnwidth]{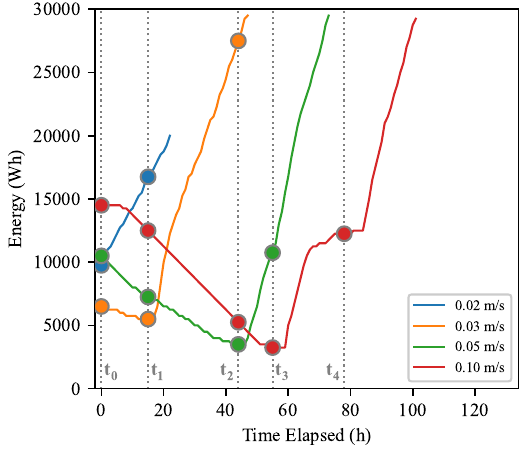}
    \caption{Overlay of minimum energy curves (retrieved from Fig.~\ref{fig:exp3_minenergy}) corresponding to a risk threshold of 10\%. The horizontal axis shows the time elapsed since UNIX timestamp 1,882,218,800 s. We identify five timestamps of interest: $t_0$ (1,882,218,800 s), $t_1$ (1,882,272,800 s), $t_2$ (1,882,377,200 s) and $t_3$ (1,882,416,800 s) are the minima of each curve, while $t_4$ (1,882,514,000 s) corresponds to the `plateau' region of the rightmost curve. The round markers indicate the start states for the Monte Carlo simulations detailed in \Cref{subsec:monte_carlo}.}
    \label{fig:exp3_startstates}
\end{figure}
Alternatively, it could deviate from the shortest (spatial) path to increase sun exposure during the traverse, and arrive at a safe haven with a sufficiently high energy level.
In the simulations presented below, the latter behavior is more prominent.

As noted in \Cref{sec:exp1}, beyond the state space discretization resolution, trial length (i.e., the number of steps taken until a terminal condition is reached) amplifies risk prediction errors.
Once again, this observation is in agreement with the theoretical convergence results derived in~\cite[Section 5]{abate_computational_2007}.
The recursive nature of dynamic programming introduces approximation errors at every step.
Given assumptions about the true risk function and the time discretization resolution, the advantage of the conservative policy is to conservatively account for these approximation errors.
A drawback of this policy, however, is that conservative choices also compound with trial length.
In \Cref{fig:exp3_minenergy}, this property appears as an increase in the minimum  energy required for a given risk threshold.
The effect is more prominent for the policy corresponding to the 0.10 m/s driving speed.
Possible ways to reduce conservative behaviour include increasing the set of actions available (e.g., allowing for multiple drive velocities or wait actions with different durations) or simply more finely discretizing the state space.
We leave these improvements as future work.

\subsection{Monte Carlo Simulations}
\label{subsec:monte_carlo}

\begin{figure*}[t]
    \centering
    \includegraphics[width=\textwidth]{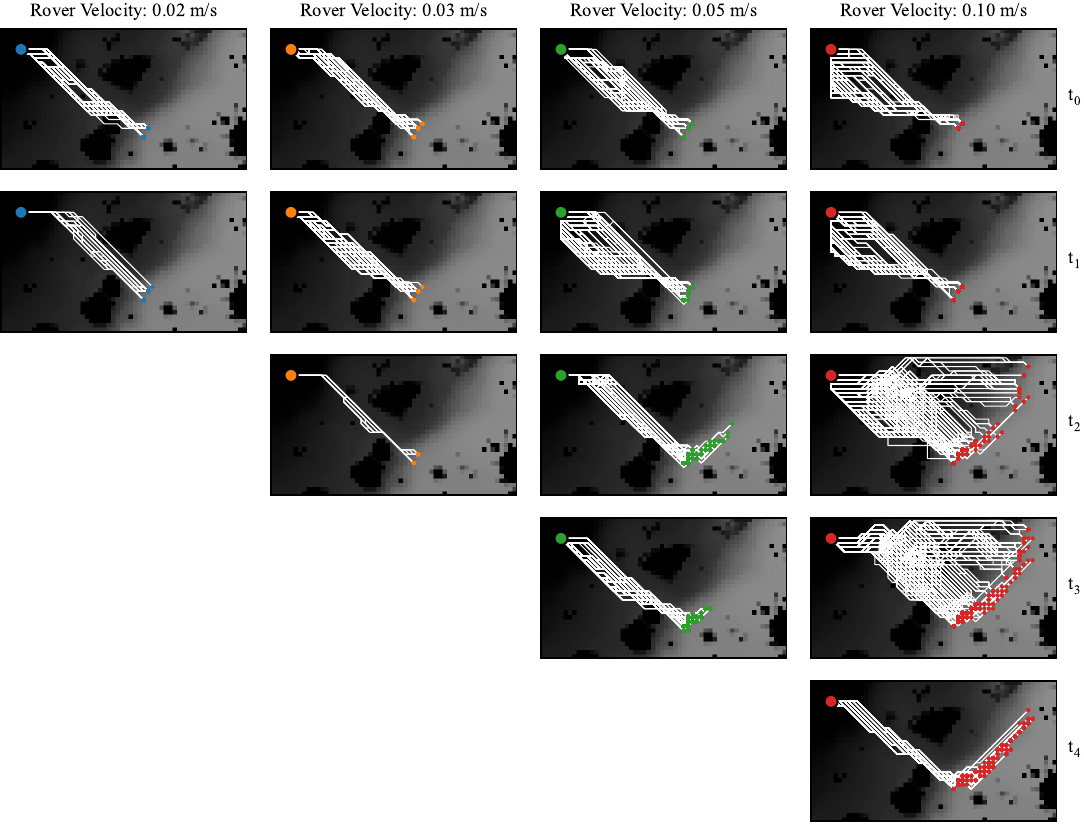}
    \caption{Overlay of successful (simulated) drives on top of a grayscale version of the average irradiance map shown in \Cref{fig:exp3_map}. Every row corresponds to a specific departure time (labelled on the right, see \Cref{fig:exp3_startstates} for details) and every column corresponds to a different recovery policy, identified by the corresponding rover drive velocity. The start location is marked with a large dot and the end locations are marked by small coloured dots.}
    \label{fig:exp3_maps}
\end{figure*}

To further highlight the differences between the four policies, we carry out and extensive series of Monte Carlo simulations.
The minimum energy curves for each policy at a risk threshold of 10\% from \Cref{fig:exp3_minenergy} are overlaid in \Cref{fig:exp3_startstates}.
The single risk threshold could, in practice, represent a hard requirement for a given  mission.
We identify five timestamps of interest: the minima of all curves are labelled with $t_0, \dots, t_3$, while $t_4$ corresponds to the `plateau' region of the curve of the policy for the greatest rover velocity.
All four policies are evaluated for departure times $t_0$ and $t_1$, three policies are evaluated for departure time $t_2$, and so on.\footnote{This does not mean that solutions do not exist for some policies for certain depature times---feasible strategies may exist with a more permissive risk threshold.}

A total of 100,000 Monte Carlo trials are carried out from each start state.
Each trial is subject to a different (random) fault profile and the same set of 100,000 fault profiles are used for each start state.
The success of individual trials and the actual risk associated with every simulation batch are calculated the same way as described in \Cref{sec:exp1}.
An overlay of traverse paths for the successful trials from each start state is shown in \Cref{fig:exp3_maps}.
Each row of plots corresponds to a different start time ($t_{0, \dots, 4}$) and every column corresponds to a different policy.
The predicted risk, the actual risk estimate, and the average length of the fault-free successful trials (i.e., number of steps taken until safety is reached) from each start state are listed in \Cref{tab:exp3_results}.

\begin{figure*}[t]
    \centering
    \includegraphics[width=\textwidth]{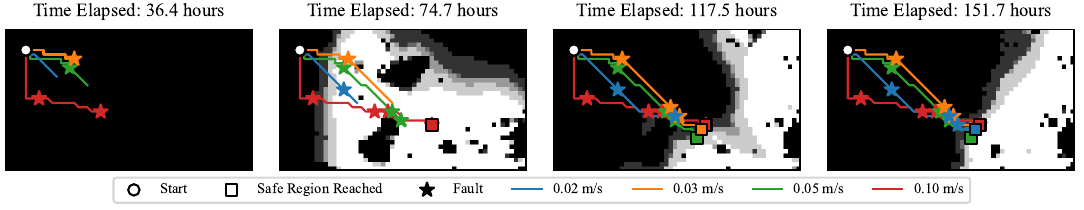}
    \caption{Four snapshots of a single simulation trial starting from time $t_0$ (refer to \Cref{fig:exp3_startstates} for the numerical value). Each shows the instantaneous solar illumination conditions over the entire ROI (black indicates total shade while white indicates full insolation) and the progress of each policy up to that time. Georeferenced map ticks are ommited for clarity but can be consulted in \Cref{fig:exp3_map}. In this trial, all four policies successfully reached safety despite being disturbed by three faults.}
    \label{fig:exp3_trial_t0}
\end{figure*}

When departing the LCROSS impact site at $t_0$, the two policies with the slowest rover velocities follow generally similar paths and head towards the nearest safe havens.
A limited velocity provides very little flexibility regarding trajectory adaptation after a fault.
Depending on the fault profile, the policy with the fastest rover velocity sometimes leads the rover south before guiding it eastward, to increase solar exposure.
Snapshots of a single simulation trial, showing the instantaneous illumination and progress with all four policies, are shown in \Cref{fig:exp3_trial_t0}.
Another contributing factor behind a larger path spread is the fixed velocity model causing longer trial lengths (and greater conservatism), as shown in the last column of \Cref{tab:exp3_results}. 

\begin{table}[t]
\centering
\begin{tblr}{
	colspec={p{0.27in} p{0.45in} p{0.6in} p{0.55in} p{0.75in}},
	row{ 1} = {rowsep=3pt},
	row{ 2} = {abovesep=3pt},
    row{15} = {belowsep=3pt},
    row{5, 9,12,14} = {belowsep=3pt},
    row{6,10,13,15} = {abovesep=4pt},
    column{2,3,4,5} = {halign = c}, 
	cell{ 2}{1} = {r = 4}{valign = m,halign = c},
	cell{ 6}{1} = {r = 4}{valign = m,halign = c},
	cell{10}{1} = {r = 3}{valign = m,halign = c},
	cell{13}{1} = {r = 2}{valign = m,halign = c},
	cell{15}{1} = {halign = c}
}
\toprule
\textbf{Start \newline Time} & 
\textbf{Velocity \newline (m/s)} & 
\textbf{Predicted \newline Risk (\%)} & 
\textbf{Actual \newline Risk (\%)} & 
\textbf{Fault Free \newline Trial Length} \\
\midrule
$t_0$  &  0.02  &  9.8  &  2.1  &  30   \\
       &  0.03  &  5.4  &  0.1  &  36   \\
       &  0.05  &  0.4  &  0.0  &  43   \\
       &  0.10  &  7.0  &  0.0  &  46   \\
\hline[dashed]
$t_1$  &  0.02  &  9.8  &  2.4  &  30   \\
       &  0.03  &  8.5  &  0.8  &  30   \\
       &  0.05  &  0.5  &  0.0  &  45   \\
       &  0.10  &  2.9  &  0.0  &  44   \\ 
\hline[dashed]
$t_2$  &  0.03  &  9.6  &  4.6  &  30   \\
       &  0.05  &  9.4  &  2.3  &  33   \\
       &  0.10  &  4.0  &  0.4  &  35   \\ 
\hline[dashed]
$t_3$  &  0.05  &  9.9  &  3.3  &  29   \\
       &  0.10  &  8.9  &  4.3  &  31   \\
\hline[dashed] 
$t_4$  &  0.10  &  9.2  &  3.3  &  31   \\
\bottomrule
\end{tblr}

\caption{Experiment 3 results summary. The first column indicates the start time label, the second column is the rover model velocity of the corresponding policy and the last column is the length of trials that were not impeded by any fault (i.e., the number of actions required to reach safety in an ideal, fault free scenario).
}
\label{tab:exp3_results}
\end{table}

\begin{figure*}[!htb]
    \centering
    \includegraphics[width=\textwidth]{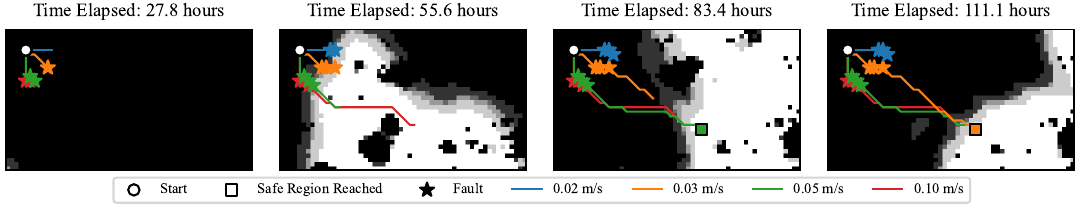}
    \caption{Four snapshots of a single simulation trial starting from time $t_1$ following the same format as \Cref{fig:exp3_trial_t0}. In this trial, three successive early faults cause the slowest traverse to barely miss a crucial solar charging period. All other policies successfully reach safety.}
    \label{fig:exp3_trial_t1}
\end{figure*}

\begin{figure*}[!htb]
    \centering
    \includegraphics[width=\textwidth]{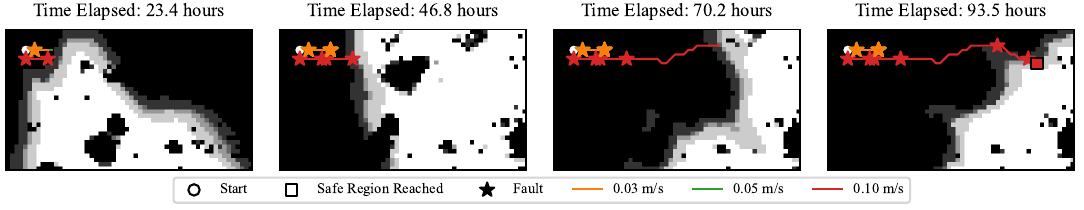}
    \caption{Four snapshots of a single simulation trial starting from time $t_2$ following the same format as \Cref{fig:exp3_trial_t0}. In this trial, three successive early faults cause all but the policy associated with the fastest rover velocity to fall too far behind schedule. The successful policy nevertheless manages to reach safety despite experiencing three additional faults.}
    \label{fig:exp3_trial_t2}
\end{figure*}

As the departure time increases, the spread in the paths of successful trials generally decreases.
Policies with a slower mobility velocity drive the rover to the nearest safe havens in an attempt to reach a safe energy level before the operational time limit.
\Cref{fig:exp3_trial_t1} shows a simulation trial departing at $t_1$ where the policy with the slowest rover velocity fails to reach safety due to three faults early into the traverse.
For the policy with the fastest rover velocity, driving down on the map from the start location has diminishing utility since the safe havens receive solar illumination sooner relative to the start time.
A departure at time $t_2$ with the fastest rover velocity illustrates drastically different successful trial paths: the rover sometimes takes advantage of the brief solar illumination period in the upper part of the map.
\Cref{fig:exp3_trial_t2} shows such an instance where only the fastest policy reaches safety.

\begin{figure*}[!htb]
    \centering
    \includegraphics[width=\textwidth]{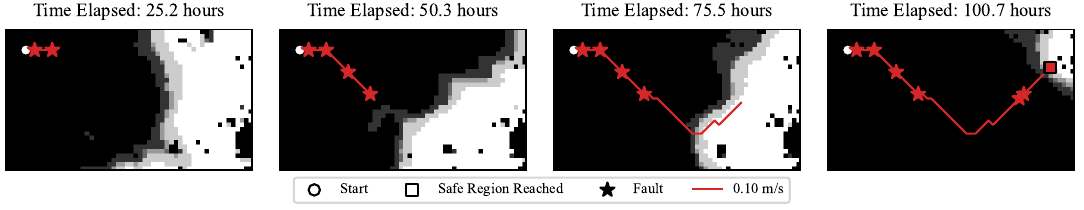}
    \caption{Four snapshots of a single simulation trial starting from time $t_4$ following the same format as \Cref{fig:exp3_trial_t0}. In this trial, the policy manages to adapt to faults occurring within the safe haven area and reach safety.}
    \label{fig:exp3_trial_t4}
\end{figure*}

The successful trials starting at time $t_4$ illustrate how fast mobility within the safe haven region allows the rover to adapt to faults occurring late in the traverse. 
This behaviour is enabled by the slower-moving sunlight in this area and explains the `plateau' region of the red curve on \Cref{fig:exp3_startstates}.
A trial where the rover drives deep into the safe haven region as it recovers from six faults is shown in \Cref{fig:exp3_trial_t4}.
Together, these simulations empirically demonstrate how higher driving velocities enable a wider range of adaptive behaviours and thus increase mission safety.

\section{Conclusion}
\label{sec:conclusion}

We have formulated a stochastic reach-avoid problem tailored to the traverse of sunlight-deprived areas at the lunar south pole by a solar-powered rover.
Our work provides a new perspective on the utility of a min-max dynamic programming approach; we conservatively account for errors caused by discretization of the state space when solving for maximally-safe policies.
We used this strategy to generate recovery strategies to exit PSRs at the lunar south pole as safely as possible.
Through the use of Monte Carlo simulations involving real terrain and insolation maps of Cabeus crater, we compared our approach against other existing  dynamic programming paradigms.
We empirically demonstrated that our method provides risk predictions that do not underestimate the true risk.
Additionally, we simulated long-range drives using orbital data from the LCROSS crash region and revealed a variety of online planning behaviours generated by assuming different rover traverse velocities.
Our results clearly show that, in addition to generating conservative recovery policies, our method enables the comparison of different rover mobility models and global mission profiles while respecting hard constraints on traverse risk.
This capability opens the door to new applications in long-range rover mission design and analysis, in-mission human operator assistance, and risk-constrained spatiotemporal planning.

There are several ways in which our work could be extended.
First, it is difficult to predict an exact spatial fault rate and fault resolution duration ahead of time, particularly in the context of planetary exploration.
Instead, it would be helpful to understand how resilient our policies are to fault profile mischaracterization.
Distributionally-robust dynamic programming methods that account for disturbance distribution uncertainty already exist~\cite{yang_dynamic_2018}, but combining the disturbance ambiguity set with ours (i.e., the state space discretization ambiguity set) would be computationally intractable for mobility problems of reasonable size.
Second, although we identified a correlation between the state space discretization resolution and the degree of conservativeness in risk prediction, it would be useful to numerically characterize this relationship. %
Third, we assumed that moving outside of the operational region of the state space immediately leads to the end of the mission.
In practice, the operational region might be defined conservatively such that a constraint violation admits the possibility of recovery.
In this context, quantifying the \textit{degree} of allowable constraint violation would be very useful.
Promising approaches might rely on risk-sensitive control~\cite{amato_how_2020, wang_risk-averse_2022}.
Lastly, combining our recovery policies with mission-level operations would enable strategic planning to optimize for scientific return while respecting mission risk constraints.

\section*{Acknowledgment}
This work was supported in part by the Natural Sciences and Engineering Research Council of Canada through the Discovery Grant program.
The work of Shantanu Malhotra was supported by Jet Propulsion Laboratory, California Institute of Technology, under a contract with the National Aeronautics and Space Administration (80NM0018D0004).
We would like to express our gratitude to Dr. Margaret P.\ Chapman for advice and insightful discussions about stochastic reachability analyses.
We would also like to thank the NASA Public Affairs Office for providing us with a high-resolution render of the VIPER rover and the NASA Jet Propulsion Laboratory, California Institute of Technology, for letting us use their lunar south pole solar illumination dataset.

\urlstyle{same}
\bibliographystyle{elsarticle-num}
\bibliography{robotics_abbrv.bib,references_clean.bib}
\end{document}